\documentclass[12pt,a4paper]{article}
\usepackage[utf8]{inputenc}
\usepackage[T1]{fontenc}
\usepackage{mathptmx}  % font

\usepackage{hyperref} % links to references, urls,...

\usepackage[margin=3cm]{geometry}  % side margin

\usepackage{graphicx}
\usepackage{placeins} % prevent images to go past barrier

\usepackage{caption}
\usepackage{subcaption}

\usepackage{mathtools}  % norm, etc.
\usepackage{amsfonts} % real number, etc

\graphicspath{{img/}} %Setting the graphicspath

\usepackage{multirow} % multi rows in tables

\usepackage[style=numeric]{biblatex} 
\begin{document}

%\title{Enhancing Saliency Maps using Ensemble Explanations}
%\title{Study of Saliency Map Fidelity using Ensemble Explanations}
%\title{Fidelity of Ensemble Aggregation for Saliency Map Explanations using Acquisition Functions from Bayesian Optimization}
\title{Fidelity of Ensemble Aggregation for Saliency Map Explanations using Bayesian Optimization Techniques}
\author{Yannik Mahlau\footnote{Corresponding Author. This work was created as part of an internship at dSPACE GmbH.}\\
Leibniz Universität Hannover\\
{\tt\small mahlau@thi.uni-hannover.de}
\and
Christian Nolde\\
dSPACE GmbH\\
%{\tt\small cnolde@dspace.de}
}
\date{July, 2022}
\maketitle

\begin{abstract}
\noindent
In recent years, an abundance of feature attribution methods for explaining neural networks have been developed. Especially in the field of computer vision, many methods for generating saliency maps providing pixel attributions exist. However, their explanations often contradict each other and it is not clear which explanation to trust. A natural solution to this problem is the aggregation of multiple explanations. We present and compare different pixel-based aggregation schemes with the goal of generating a new explanation, whose fidelity to the model's decision is higher than each individual explanation. Using methods from the field of Bayesian Optimization, we incorporate the variance between the individual explanations into the aggregation process. Additionally, we analyze the effect of multiple normalization techniques on ensemble aggregation. 
\end{abstract}

\section{Introduction}
Neural Networks in computer vision applications, like the popular ResNet18 \cite{he16} architecture, often have millions of parameters and are therefore inherently difficult to interpret. This is a big deterrent for the use of neural networks in real world applications \cite{molnar22}, as understanding is vital for creating trust in the machine learning model \cite{miller19}. Therefore, the topic of explainable machine learning has drawn a lot of attention in the research community \cite{ras20}. Even though we focus on the application of neural networks in computer vision tasks, the same is true for all fields of machine learning involving neural nets. Joshi et al.\ \cite{joshi21} provide an overview of explainability in different machine learning fields. As a result of this interest, a lot of different explanation schemes have been developed. However, their application in real-world tasks is challenging, because it is not clear which explanation represents the models decision best. The  quality of an explanation regarding the models decision is called fidelity \cite{molnar22}.

A comparison between explanations is problematic, because no ground truth data exists for the models decision process. Additionally, visual comparison by humans has been proven to be precarious \cite{adebayo18} since humans inherently have a confirmation bias of their own belief. As a result, various metrics for measuring fidelity have been developed \cite{zhou21}. Most commonly used are the insertion and deletion metric \cite{petsiuk18}, even though valid criticism regarding the issue of baseline choice and implicit independence assumptions can be voiced \cite{wang22}. 

Instead of developing more algorithms for generating individual explanation, we believe it is pertinent to use multiple explanations and aggregate them into a new and better explanation. For the aggregation of multiple explanations, we use acquisition functions from the field of Bayesian Optimization as they are a tool for estimating the importance of a data point (pixel). In contrast to Bayesian Optimization, we utilize acquisition functions for identifying data points with low disagreement (low standard deviation). That is, because a pixel is most important if all individual explanations agree on its importance. One further difference to Bayesian Optimization is the absence of a sequential process, because the individual explanations yield fixed data. Therefore, we only evaluate the acquisition function once for every pixel.

The main contributions of our work are:
\begin{itemize}
    \item We introduce different aggregation functions for ensemble explanations. Empirically, we evaluate their fidelity on two different datasets.
    \item For aggregation methods with adjustable hyperparameters, we test their sensitivity.
    \item We compare different normalization techniques and empirically stress their importance.
\end{itemize}

\section{Related Work}
In recent years, a number of pixel-attribution (saliency map) methods for explaining computer vision classification models have been developed. They can be broadly categorized into model-agnostic (black-box) and model-specific (white-box) methods \cite{molnar22}.

Model-agnostic interpretation schemes do not consider the internal structure of a model, but only the relation of input and output. In RISE \cite{petsiuk18}, for example, part of the input image is occluded and the resulting change in output probability is observed.

In contrast, model-specific methods make use of the internal activations, weights and gradients of a model. Therefore, these methods have more information available, but are also not universally applicable anymore. GradCAM \cite{selvaraju19}, for instance, linearly weighs the activations of the last convolutional layers with the gradient of fully connected layers. In the extension PolyCAM \cite{englebert22}, all activations of convolutional layers are aggregated instead of just the last layer. SmoothGrad \cite{smilkov17} follows a different idea by averaging the gradients of the model output with regard to input images perturbed by small random noise. Integrated Gradients \cite{sundararajan17} does not only consider a single gradient per image, but rather integrates all gradients along an interpolation path from a baseline image to the original. Integrated Gradients and SmoothGrad can be combined to form Expected Gradients \cite{erion19}. Integrated Gradient Optimized Saliency (iGOS++) \cite{khorram20} utilizes the integrated gradients to optimize a soft mask using stochastic gradient descent.

\begin{figure}[ht]
    \centering
    \includegraphics[width=\textwidth]{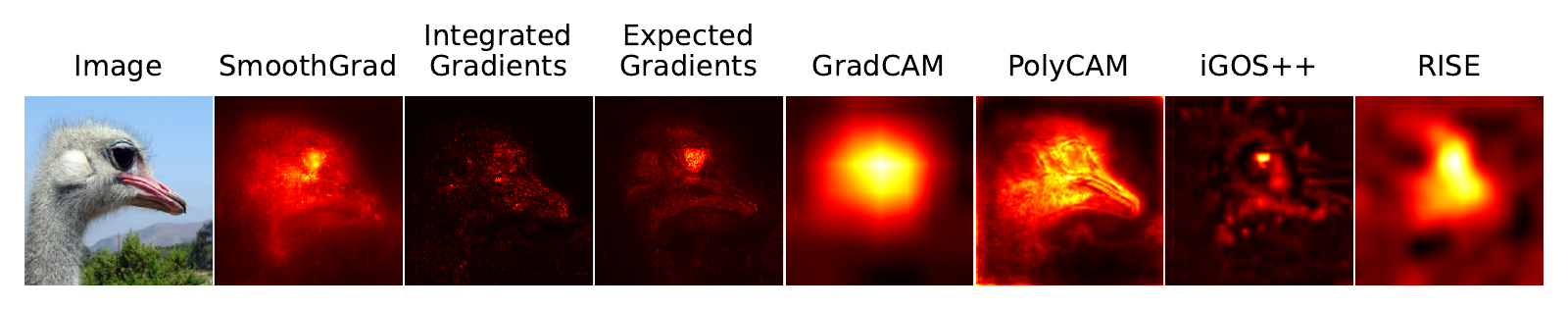}
    \caption{Comparison of base attribution methods for ResNet18 trained on ImageNet. Attributions are linearly normalized to the full color range.}
    \label{fig:base_attr}
\end{figure}

In Figure \ref{fig:base_attr}, we display the attributions of all mentioned methods on an example image. We normalize the attributions to the full color range for better visualization in all figures throughout this work. A common observation comparing different saliency methods is that they disagree with each other. Krishna et al.\ \cite{krishna22} developed multiple metrics measuring this disagreement. The question arises, if a combination of multiple methods may produce a better saliency map than each individual one. Bhatt et al.\ \cite{bhatt20} developed aggregation schemes minimizing explanation sensitivity by using Shapley Values. Additionally, they  minimized explanation complexity by gradient-descent and region shrinking.

Rieger et al.\ \cite{rieger20} proved that an average of multiple attribution methods is more robust to adversarial attacks than a single explanation. Incorporating confidence, this method can also be used to enhance fidelity \cite{rieger19}. Correspondingly, a Restricted Boltzmann Machine \cite{smolensky86} trained on an ensemble of explanations, using unsupervised contrastive divergence learning \cite{hinton06}, is robust against noisy and unfaithful explanations \cite{borisov21}. Additionally, it is possible to achieve high fidelity using this ensemble aggregation. 

\section{Method}
In the context of computer vision, we consider a machine learning model for classification $F: \mathbb{R}^{m \times n \times d} \rightarrow \mathbb{R}^{|C|}$, which maps an input image $I \in \mathbb{R}^{m \times n \times d}$ to a probability vector  representing the prediction for each class $c \in C$. The input image has a spatial resolution of $m\times n$ pixels with $d$ color channels.

An explanation of a single prediction $e(F, I, c) \in\mathbb{R}^{m \times n}$ yields an attribution score $e_{i,j}$ for every pixel $(i,j)$ of the input image. For better readability, we omit the parameters $F$, $I$ and $c$. An ensemble $E = \{e^1, ..., e^k\}$ is a set of attributions. We generate the members $e^i$ of an ensemble using different attribution methods and call them base attributions. The base attributions of an ensemble can be merged into a new explanation $\hat{e} \in \mathbb{R}^{m \times n}$ using an aggregation function $g: E \mapsto \hat{e}$. We present multiple aggregation functions in Section \ref{sec:agg}.

Note that our methods could easily be extended to object detection. However, we only consider image classification as there are many more base attribution methods available.

\subsection{Normalization}
Normalization of the base attributions is necessary, because they do not share the same scale. For example, the pixel attributions of RISE represent a probability difference in the interval $[0, 1]$, but the attributions of PolyCAM are a product of activations, which can take arbitrary (real) positive values. Conversely, the gradients of SmoothGrad can theoretically be any real number, but are practically often small values around zero. If one would aggregate an ensemble without normalization, the high attribution values of PolyCAM would dominate the small values of SmoothGrad.
Images are often linearly mapped to the interval $[0, 1]$ to utilize the full color or greyscale range \cite{gonzales07}. Equivalently, we can compute a normalized attribution (see Equation \ref{eq:lin}).
Another normalization technique is Z-score normalization \cite{kreyszig12}. Here, we assume that the  distribution of the attribution is Gaussian with mean $\mu(e)$ and standard deviation $\sigma(e)$ over the pixels values. This distribution can then be normalized to a standard normal distribution by Equation \ref{eq:z}.
Additionally, we experimented with the $l_1$- and $l_2$-normalization (see Equation \ref{eq:norm1} and \ref{eq:norm2}).
This gives us the following normalization methods:

\parbox[ht]{0.4\linewidth}{%
\begin{align}
\tilde{e}_{lin} &= \frac{e - \min(e)}{\max(e) - \min(e)}, \label{eq:lin}  \\
\tilde{e}_z &= \frac{e - \mu(e)}{\sigma(e)}, \label{eq:z}
\end{align}
}\hfill
\parbox[ht]{0.4\linewidth}{
\begin{align}
\tilde{e}_{1} &= \frac{e}{\lVert e \rVert_1}, \label{eq:norm1} \\
\tilde{e}_{2} &= \frac{e}{\lVert e \rVert_2}. \label{eq:norm2}
\end{align}
}

\begin{figure}[ht]
    \centering
    \includegraphics[width=0.75\textwidth]{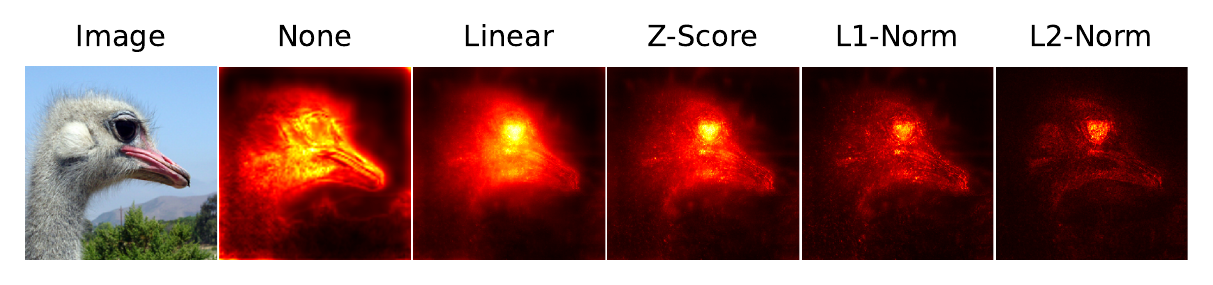}
    \caption{Comparison of normalization techniques on average aggregation. For the base attributions, see Figure \ref{fig:base_attr}. The explained model is a ResNet18 trained on ImageNet. Attributions are linearly normalized to the full color range after aggregation.}
    \label{fig:norm_comp}
\end{figure}

Figure \ref{fig:norm_comp} displays the average attribution (i.e. mean of the normalized base attributions) using the presented normalization techniques. See Equation \ref{eq:avg} for more details on average aggregation. As in all figures, the displayed aggregated attributions are linearly normalized to the full color range for visualization purposes. 
Without normalization, PolyCAM dominates all other base attributions of the ensemble (see Figure \ref{fig:base_attr} for comparison). This effect diminishes when using linear normalization. However, base attributions with many high pixel attributions like GradCAM or RISE contribute more. That is, because in linear normalization only the minimum and maximum values are used. In contrast, Z-score-, $l_1$- and $l_2$-normalization prohibit this issue by considering the full distribution for normalization. We perform an empirical comparison of all mentioned normalization techniques in Section \ref{sec:exp_norm}. 

\subsection{Aggregation}
\label{sec:agg}
We present a number of different aggregation functions for ensemble explanations. The simplest is the average of base attributions:
\begin{equation}
    g_{avg}\left(E\right) = 
        \frac{1}{|E|} 
        \sum_{e \in E}{e}. 
    \label{eq:avg}
\end{equation}

A different, non-linear aggregation method is the median, or a general percentile. We denote this aggregation as $g_k(E)$, where $k\in\mathbb{R},\; 0\leq k \leq 100$ is the percentage of the percentile. For example, $g_{50}(E)$ is the median aggregation and $g_{25}(E)$ the lower quartile. We use linear interpolation between data points, but other methods like nearest neighbor or midpoint are also viable.

Additionally, it might be helpful to balance the mean and standard deviation of base attributions. If a pixel is important, the mean attribution of the pixel should be high and the standard deviation low. The mean attribution $\mu(E_{i,j})$ of a pixel $(i,j)$ and its standard deviation $\sigma(E_{i,j})$ can be calculated by:
\begin{align}
    \mu\left(E_{i,j}\right) &=
    g_{avg}\left(E_{i,j}\right) = 
        \frac{1}{|E|} 
        \sum_{e\in E} 
            e_{i,j}, \\
    \sigma\left(E_{i,j}\right) &= 
        \sqrt{
            \frac{1}{|E|} 
            \sum_{e\in E} 
            \left(
                e_{i,j} - \mu \left( E_{i,j} \right) 
            \right)^2 }.
\end{align}

Note that this is a mean and standard deviation over multiple attributions for a single pixel. In contrast, Z-score normalization uses the global mean and standard deviation over all pixels of a single attribution.

Methods balancing mean and standard deviations can be found by the name "acquisition function" in the field of Bayesian Optimization \cite{agnihotri20}. They assign a score to every data point, which indicates the value of a sample at this position in the search space. Our motivation for using acquisition functions is that this value transfers to the pixel importance when applied to an ensemble of saliency maps.

Three commonly used acquisition functions are Upper Confidence Bounds (UCB) \cite{auer03}, Probability of Improvement (PI) \cite{kushner64} and Expected Improvement (EI) \cite{jones98}. They balance mean and standard deviation with the exploration rate $\epsilon$. The name "exploration rate" originates from the setting of Bayesian Optimization, where it balances between exploration and exploitation in the search space. They are calculated as:
\begin{align}
    g_{ucb}\left(E_{i,j}, \epsilon\right) &=
    \mu\left(E_{i,j}\right) 
    + \varepsilon 
    \sigma\left(E_{i,j}\right), \\
    g_{pi}\left(E_{i,j}, \epsilon\right) &=
    \Phi\left(Z\right),  \\
    g_{ei}\left(E_{i,j}, \epsilon\right) &=
    \Phi\left(Z\right) \left(
        \mu\left(E_{i,j}\right) 
        - E^+ - \epsilon
    \right) 
    + \sigma\left(E_{i,j}\right) \phi\left(Z\right) .
    \label{eq:ei}
\end{align}

$Z = \frac{\mu(E_{i,j}) - E^+ - \epsilon}{\sigma(E_{i,j})}$ is a random variable and $E^+ = \max_{i,j}\left(\mu\left(E_{i,j}\right)\right)$ is the approximate attribution value of the most important pixel. 
$\phi$ denotes the probability distribution function of a standard normal distribution and $\Phi$ the respective cumulative distribution function. 
All three aggregation functions are highly dependent on the exploration rate $\epsilon$. However, it is often not clear what exploration rate leads to the best result. To alleviate this issue, Jasrasaria et al.\ \cite{jasrasaria18} propose to use 
\begin{equation}
    \epsilon = \frac{1}{m n E^+} \sum_{i=1}^{m} \sum_{j=1}^{n} \sigma(E_{i,j})^2
\end{equation}
in Equation \ref{eq:ei}. They call their method Contextual Improvement (CI), which we denote by the function $g_{ci}(E)$.

Additionally, we propose to use the average attribution for different exploration rates. Naturally, this is only viable for PI and EI. For UCB, this method would simply lead to the same result as using the average exploration rate, due to the linear nature of UCB. For an interval $[a,b]\subseteq \mathbb{R}$ we define Average Probability of Improvement (API) and Average Expected Improvement (AEI) of an ensemble as:
\begin{align}
     g_{api}\left(E_{i,j}, a, b\right) &=
        \frac{1}{b-a}
        \int_{a}^{b} g_{pi}(E_{i,j}, \epsilon) \; \mathrm{d}\epsilon, \\
     g_{aei}\left(E_{i,j}, a, b\right) &= 
     \frac{1}{b-a}\int_{a}^{b} g_{ei}(E_{i,j}, \epsilon) \; \mathrm{d}\epsilon.
\end{align}
We approximate the average using $N$ discrete samples evenly spaced in the interval $[a,b]$:
\begin{align}
     g_{api}\left(E_{i,j}, a, b\right) &\approx 
     \frac{1}{N}\sum_{k=0}^{N-1} 
    g_{pi}\left(
        E_{i,j}, \; \frac{k(b-a)}{N-1}+a 
    \right), \\
     g_{aei}\left(E_{i,j}, a, b\right) &\approx
     \frac{1}{N}\sum_{k=0}^{N-1} 
     g_{ei}\left(
        E_{i,j}, \; \frac{k(b-a)}{N-1}+a 
    \right).
\end{align}

Lastly, we also consider the aggregation functions presented in earlier literature. The aggregation using a Restricted Boltzmann Machine (RBM) \cite{borisov21} is denoted by $g_{rbm}(E_{i,j}, \alpha, k)$ with learning rate $\alpha\in\mathbb{R}$ and $k\in\mathbb{N}$ iterations of training in the contrastive learning algorithm. 

Furthermore, we also consider the aggregation introduced by Rieger et al.\ \cite{rieger19}, which they denote Variance-Aggregation (VAR). We extend their concept by the parameter $\epsilon$ controlling the balance between mean and standard deviation:
\begin{align}
     g_{var}\left(E_{i,j}, \epsilon, \delta\right) = 
     \frac{1}{|E|} \sum_{e \in E} 
        \frac{e_{i,j}}{
            \epsilon \sigma\left(E_{i,j}\right) + \delta
        }.
\end{align}

\begin{figure}[ht]
    \centering
    \includegraphics[width=\textwidth]{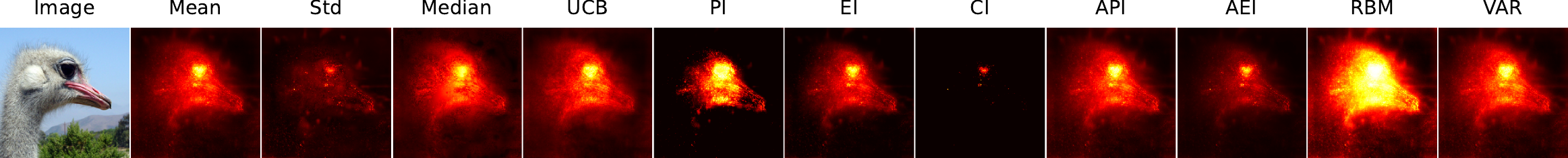}
    \caption{Comparison of Aggregation Methods. The explained model is a ResNet18 trained on ImageNet. The base attributions are normalized using Z-score normalization. After aggregation, the results are linearly normalized to the full color range.}
    \label{fig:aggr_comp}
\end{figure}

In Figure \ref{fig:aggr_comp}, all mentioned aggregation functions are shown. The attributions are linearly normalized to the full color range as in all other figures. The mean and median saliency map inherently produce very similar results. UCB includes the standard deviation such that more details are visible. PI and CI yield sparse masks displaying only the most important pixels, for example the eye of the ostrich. The attribution of EI and AEI separate important and non-important regions, e.g.\ the edge between the ostrich and the background. The saliency map of API is similar, but more scattered. RBM produces a saliency map with approximated binary values. The explanation of VAR is intrinsically close to the average aggregation. The results of methods using hyperparameters (exploration rate, ...) are highly dependent on these hyperparameters. For more details on the effect of hyperparameters, see Section \ref{subsec:hp}.

\section{Experiments}
\label{sec:experiments}
We perform different experiments to evaluate the theory presented. To measure the quality of an explanation, we utilize the insertion and deletion metrics \cite{petsiuk18}. The idea of the insertion metric is to incrementally insert the most important pixels into a baseline image. Similarly, the deletion metric replaces the most important pixels from the original image with the baseline. For every step, the model output of the class of interest is measured. A good explanation results in a steeply rising insertion and steeply falling deletion curve. Therefore, the area under the curve (AUC) should be maximized in insertion and minimized in the deletion metric. We use 1000 increments in our experiments. Therefore, in the ImageNet-dataset about 64 pixels are inserted or deleted each increment. For GTSRB, the size of an increment is about 16 pixels.

The choice of baseline is generally important for many feature attribution methods and metrics \cite{sturmfels20}. On the one hand it has to contain no information, i.e. the prediction of the model changes in comparison to the original. On the other hand, it cannot be too far from the training distribution, i.e. an out of distribution sample. We employ a random normal baseline as a solution. More precisely, we estimate mean and standard deviation of each color channel and sample new values for every pixel.

There exist many metrics similar to the insertion and deletion metric, namely IROF \cite{rieger20irof}, pixel flipping \cite{samek15}, masking top pixels \cite{chen18}, infidelity \cite{yeh19} and many more not listed here. For simplicity, we do not consider these.

Another popular metric measuring explanation fidelity is the pointing game \cite{zhang16}. However, as discussed by Khorram et al.\ \cite{khorram20}, it relies on human-annotated ground truth boxes. Therefore, the metric does not measure the models decision process well. We do not consider the pointing game for this reason.

For all experiments, the ensemble consists of the following base attributions. For the detailed hyperparameter configuration refer to Section \ref{sec:hp_conf} in the appendix. 
\vspace{-.2cm}
\begin{itemize}
\setlength\itemsep{-0.1cm}
    \item SmoothGrad \cite{smilkov17} 
    \item Integrated Gradients \cite{sundararajan17}
    \item Expected Gradients \cite{erion19}
    \item GradCAM \cite{selvaraju19}
    \item PolyCAM \cite{englebert22}
    \item iGOS++ \cite{khorram20}
    \item RISE \cite{petsiuk18}
\end{itemize}

The experiments were performed on a Intel(R) Xeon(R) W-2145 CPU (3.70GHz, 8 Core, 16 Threads), Nivida RTX 6000 GPU and 32GB RAM. We average our results over 2000 sample images from the respective validation sets. The methods were implemented using TensorFlow 2.7.1 \cite{abadi15} and Scikit-learn 1.0.2 \cite{pedregosa11}.

\subsection{Aggregation Comparison}
We compare the different ensemble aggregations on two different datasets: the ImageNet-dataset \cite{deng09} and the German Traffic Sign Recognition Benchmark (GTSRB) \cite{stallkamp12}. The model architecture is a ResNet18 \cite{he16}. The ImageNet-dataset contains over 14 million images from 1000 classes with a resolution of $224\times224$ pixels. The GTSRB-dataset contains more than 50000 images from 43 classes with a resolution of $128\times128$ pixels.

\begin{figure}[ht]
\centering
\begin{subfigure}{.5\textwidth}
  \centering
  \includegraphics[width=\linewidth]{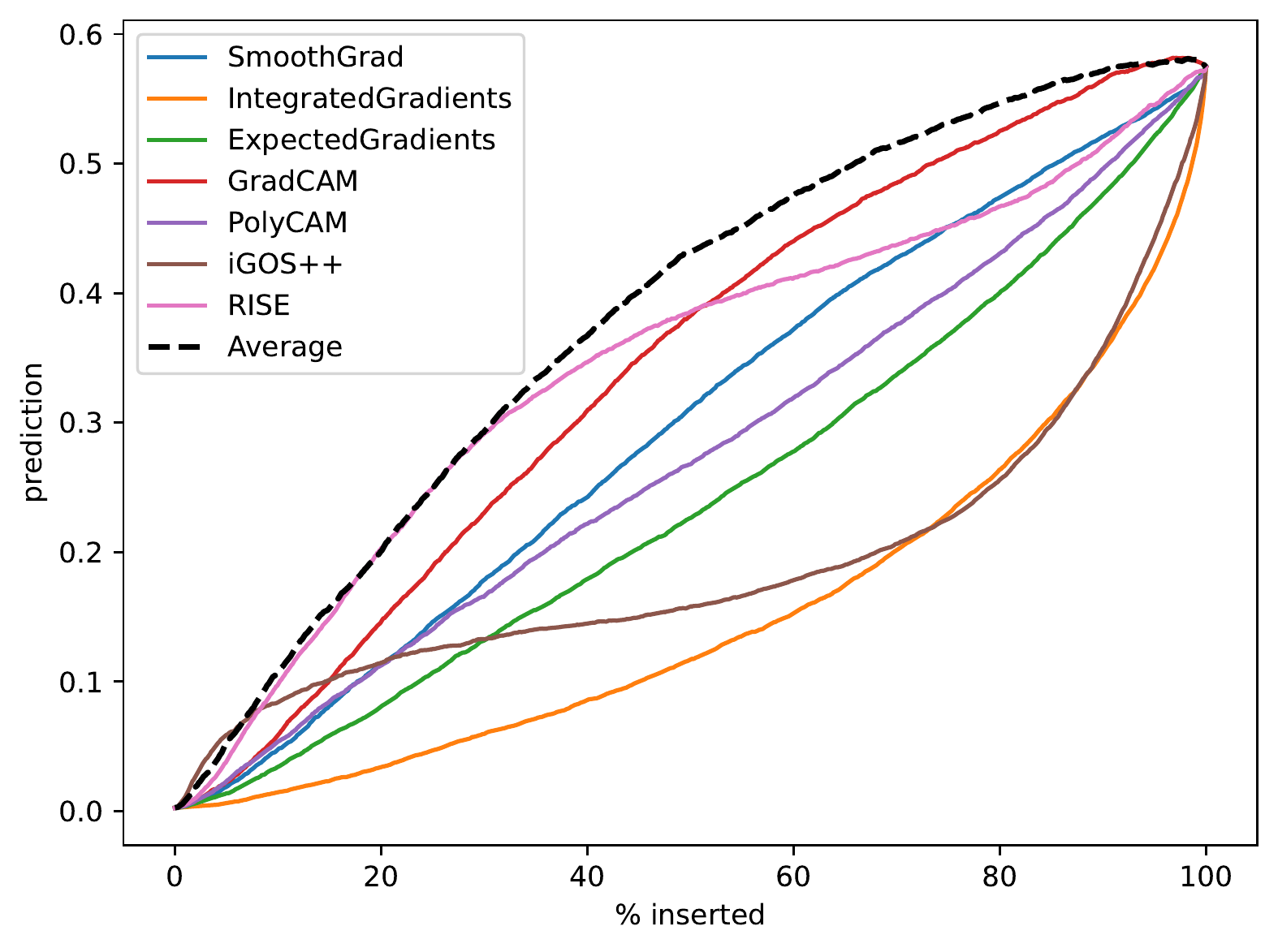}
  \caption{Insertion Curves}
  \label{fig:base_ins_curve}
\end{subfigure}%
\begin{subfigure}{.5\textwidth}
  \centering
  \includegraphics[width=\linewidth]{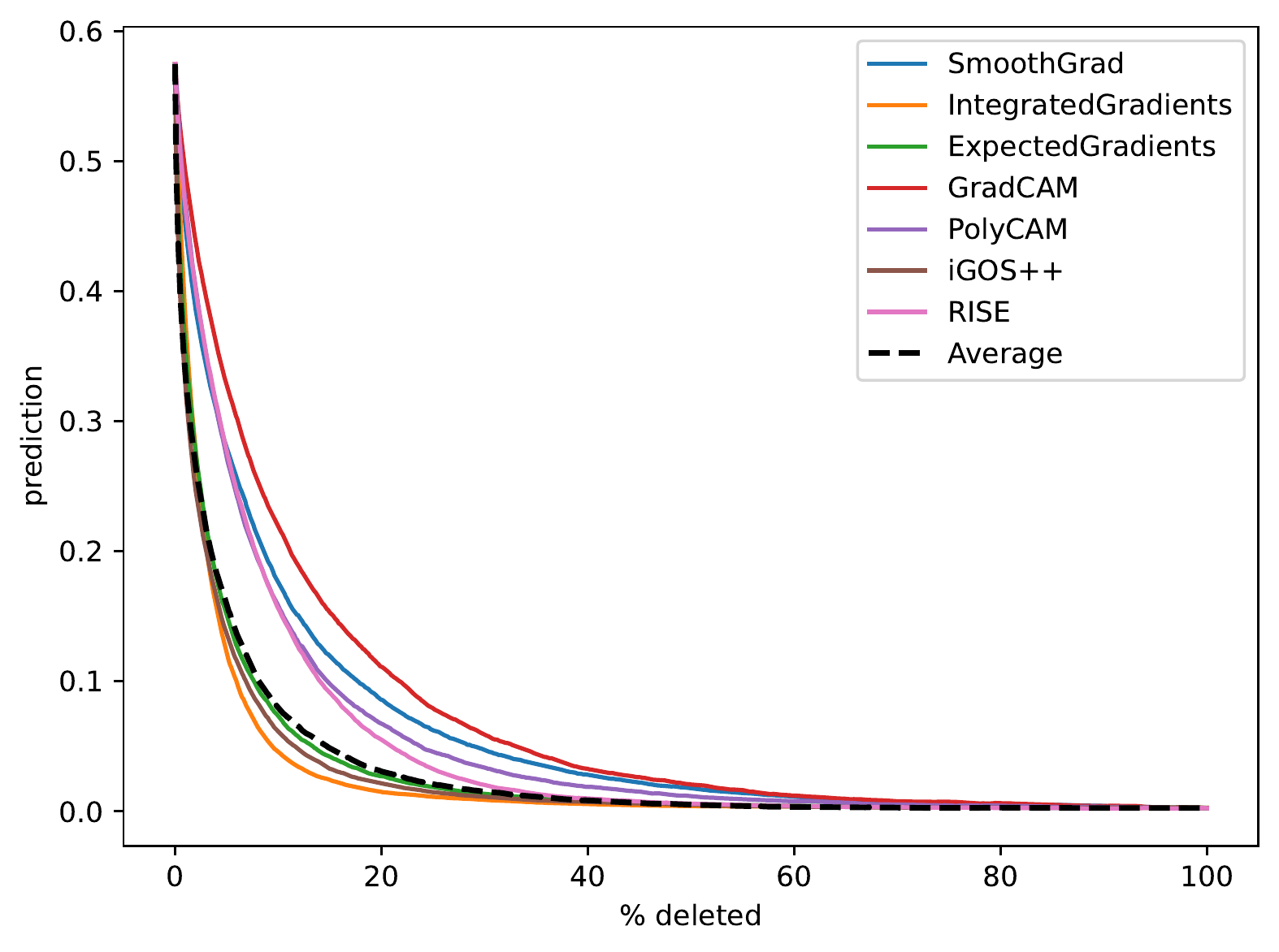}
  \caption{Deletion Curves}
  \label{fig:base_del_curve}
\end{subfigure}
\caption{Insertion and deletion curves for base attributions and average aggregation. The explained model is a ResNet18 trained on the ImageNet-dataset. Results are averaged over 2000 sample images.}
\label{fig:base_curve}
\end{figure}

In Figure \ref{fig:base_curve}, we display the insertion and deletion curves for the first experimental setup using a ResNet-18 trained on the ImageNet dataset. The average attribution achieves a better insertion score than each individual base attribution. It utilizes the initial steep ascend of iGOS++ as well as the continuous ascend of RISE and GradCAM. Note that the S-shaped curve of iGOS++ results from the sparse saliency map. iGOS++ assigns an attribution score of zero to many pixels, which are in turn inserted or deleted in random order. 

The deletion score for the average attribution is not better than all base attributions, but close to the best. Additionally, the base attributions which achieve a better deletion score, namely Integrated Gradients, Expected Gradients and iGOS++, perform much worse overall in the insertion metric. Overall we can conclude that the average attribution is better than each individual base attribution.

\begin{table}[htb]
\centering
\begin{tabular}{|l|c|c|c|c|}
    \hline
    Aggregation Method & \multicolumn{2}{c|}{ImageNet} & \multicolumn{2}{c|}{GTSRB} \\
                       & Insertion & Deletion & Insertion & Deletion \\
    \hline
    Average               & 0.3808          & 0.0296          & 0.6740          & 0.0667 \\
    RBM $(\alpha=10^{-4}, k=100)$& 0.3827   & 0.0301          & 0.6312          & 0.0744 \\
    VAR $(\epsilon=10^{-7})$& 0.3819        & 0.0296          & 0.6739          & 0.0671 \\
    UCB $(\epsilon=-0.5)$ & \textbf{0.3896} & 0.0298          & \textbf{0.6798} & \textbf{0.0657} \\
    Percentile $(k=25)$   & 0.3713          & \textbf{0.0295} & 0.6720          & 0.0675 \\
    PI $(\epsilon=-4)$    & 0.3098          & 0.0304          & 0.6306          & 0.0817 \\
    EI $(\epsilon=-6)$    & 0.3397          & 0.0298          & 0.6681          & 0.0673 \\
    CI                    & 0.2646          & 0.0324          & 0.5908          & 0.0827 \\
    API $([a,b]=[-7, 2])$ & 0.3546          & 0.0299          & 0.6726          & 0.0672 \\
    AEI $([a,b]=[-10, 5])$& 0.3692          & 0.0298          & 0.6721          & 0.0672 \\
    \hline
\end{tabular}
\caption{Performance of Ensemble Aggregation Methods. Base Attributions are normalized using Z-score normalization. Hyperparameters were tuned on ImageNet and kept constant for GTSRB. Results are averaged over 2000 samples.}
\label{tab:agg_res}
\end{table}

In Table \ref{tab:agg_res}, we list the performance of all aggregation functions on both datasets. We tuned the hyperparameters of all methods on the ImageNet dataset and used the same values for GTSRB. This tests the general applicability of the aggregations.

Across both datasets, the simple average aggregation achieved surprisingly high performance. Many methods were not able to achieve higher scores in any metric on any dataset. Most drastically, CI performed the worst in all experiments. That is, because the approximated best value for the exploration rate $\epsilon$ in Bayesian Optimization does not transfer to explainability. In the context of Bayesian Optimization, a sparse score distribution is desirable, but this directly leads to a bad insertion and deletion score. PI fails for the same reason, even though not as drastically. EI, API and AEI are able to achieve scores nearly as good as the average attribution, but always a little worse. Therefore, we deem CI, PI, EI, API and AEI not applicable in the context of explainability ensembles. 

RBM achieves a higher insertion score than average aggregation on the ImageNet dataset. However, the hyperparameter configuration did not transfer well to GTSRB, where the insertion score is much lower. In general, it was very challenging to find good hyperparameters for learning rate $\alpha$ and number of iterations $k$. Borisov et al.\ \cite{borisov21} reported deletion scores significantly lower than average aggregation, but we were not able to reproduce these results using our models and datasets.

Variance aggregation (VAR) has a better insertion score than average aggregation on ImageNet, but this small advantage diminishes on GTSRB. 

The lower quartile aggregation (percentile with $k=25$) was able to achieve consistent good scores and even managed to beat the average in the deletion metric on ImageNet. However, the differences in the deletion scores on ImageNet are so small that we do not judge the lower quartile to be better than average aggregation overall.

The best performance was achieved by UCB. It earned the best score on three experiments and was very close on the fourth one. Admittedly, the difference between average and UCB is small, but the hyperparameter configuration of $\epsilon=-0.5$ seems to transfer well to other datasets. Therefore, we believe that incorporating the uncertainty through UCB is an uncomplicated method of enhancing the average aggregation.

\subsection{Hyperparameter Study}
\label{subsec:hp}
We compare the hyperparameters of percentile aggregation, UCB, PI, EI and variance aggregation. All experiments presented in this section are performed using a ResNet18 trained on the ImageNet-dataset.

\subsubsection{Percentile Aggregation}

\begin{figure}[ht]
\centering
\begin{subfigure}{.5\textwidth}
  \centering
  \caption{Insertion}
  \includegraphics[width=\linewidth]{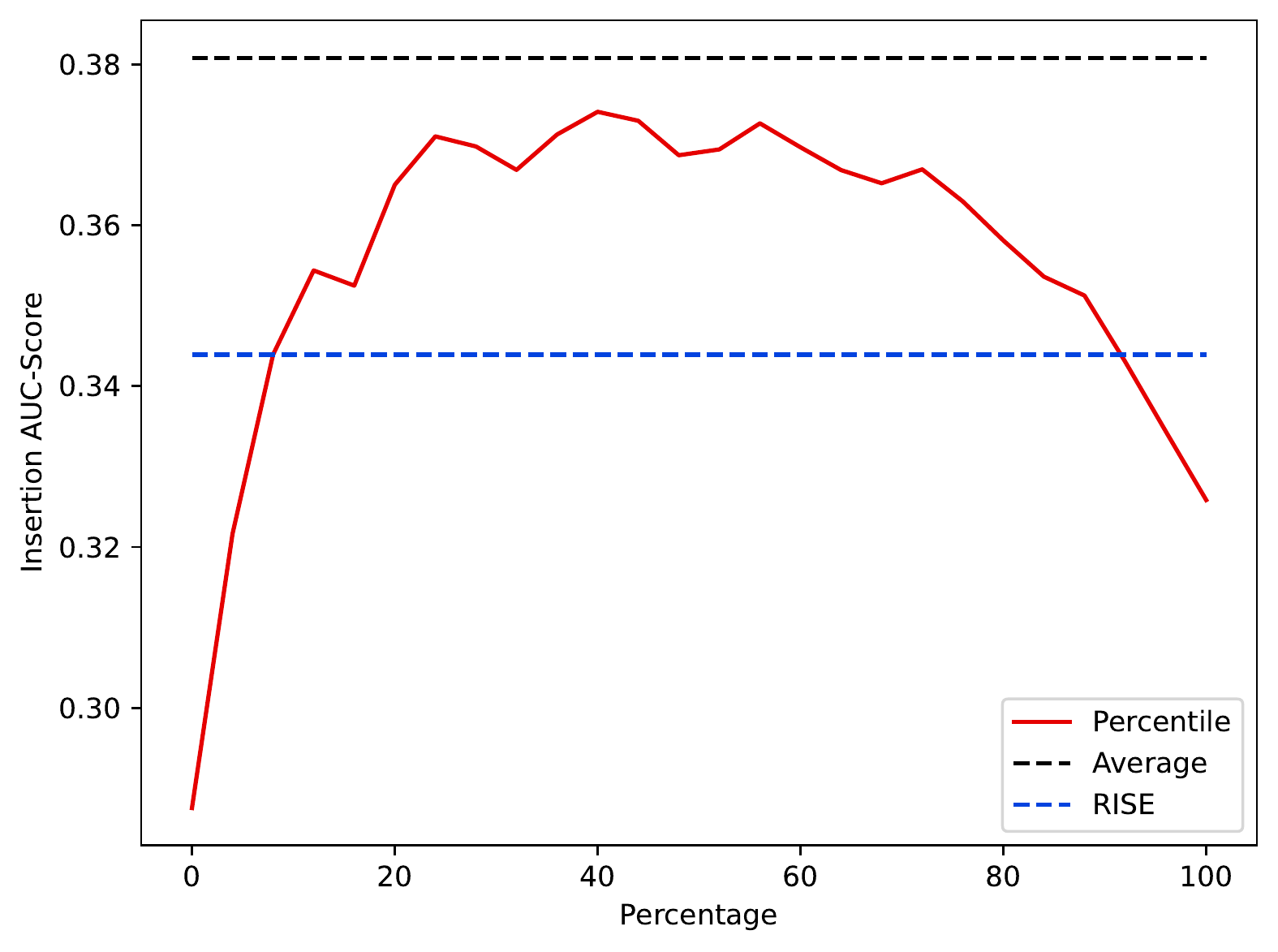}
  \label{fig:perc_ins}
\end{subfigure}%
\begin{subfigure}{.5\textwidth}
  \centering
  \caption{Deletion}
  \includegraphics[width=\linewidth]{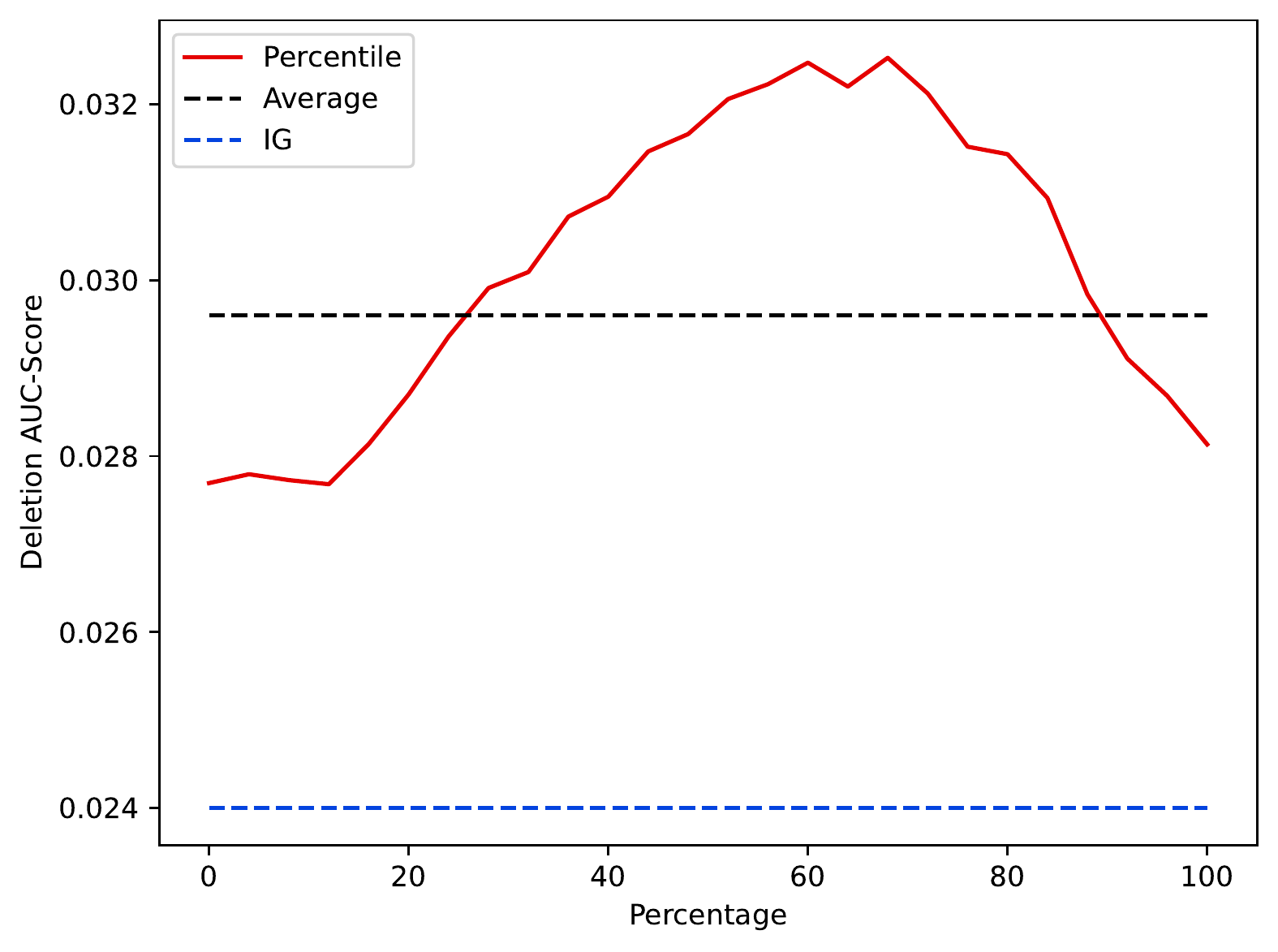}
  \label{fig:perc_del}
\end{subfigure}
\begin{subfigure}{\textwidth}
  \centering
  \vspace{-.3cm}
  \includegraphics[width=0.9\textwidth]{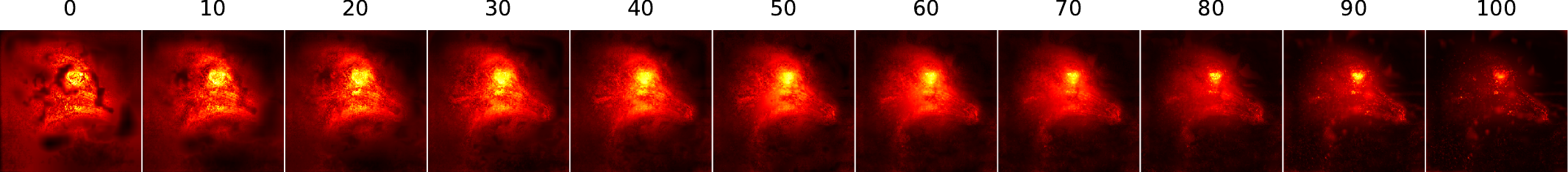}
\end{subfigure}
\caption{Insertion and deletion scores for different percentages of percentile aggregation. Base attributions are normalized using Z-score. The explained model is a ResNet18 trained on ImageNet. Results are averaged over 2000 samples. }
\label{fig:perc_hp}
\end{figure}

The percentile aggregation $g_k(E)$ has the percentage parameter $k$. We compare the insertion and deletion score for different values of k. The aggregation with $k=0$ corresponds to the minimum, $k=50$ to the median and $k=100$ to the maximum value. Figure \ref{fig:perc_hp} displays the saliency maps for different values of $k$ as well as their insertion and deletion scores. The best insertion score is achieved at $k=40$ and the best deletion score at $k=10$. Interestingly, a good insertion score seems to lead to a bad deletion score. 

Percentile aggregation manages to achieve a higher insertion score than the best base attributions over a large range of $k$. However, it is never able to beat the average aggregation. The order of magnitude is much smaller for the deletion score (note the different y-axis-scale).

\subsubsection{Upper Confidence Bound}

\begin{figure}[ht]
\centering
\begin{subfigure}{.5\textwidth}
  \centering
  \caption{Insertion AUC}
  \includegraphics[width=\linewidth]{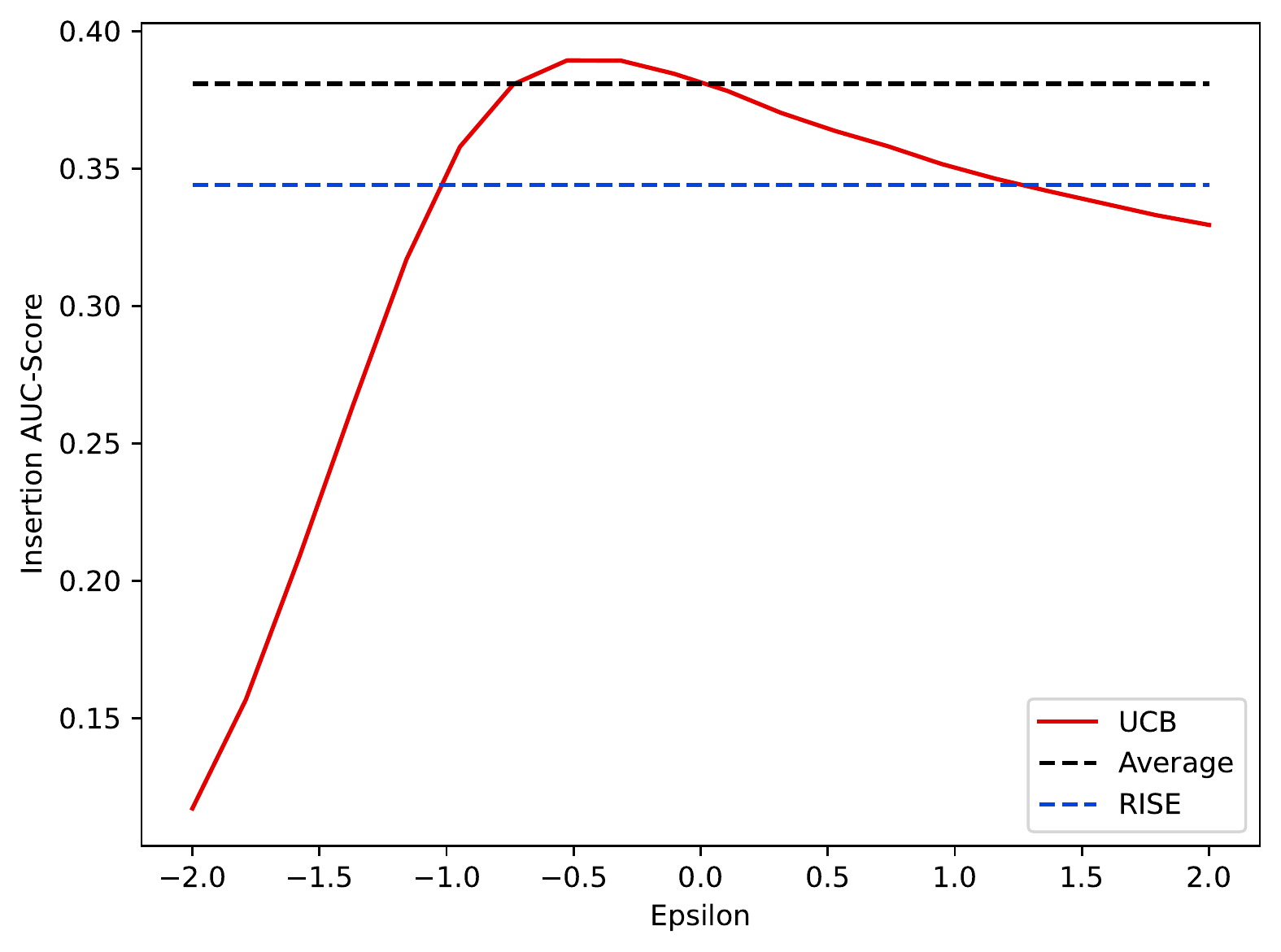}
  \label{fig:ucb_ins}
\end{subfigure}%
\begin{subfigure}{.5\textwidth}
  \centering
  \caption{Deletion AUC}
  \includegraphics[width=\linewidth]{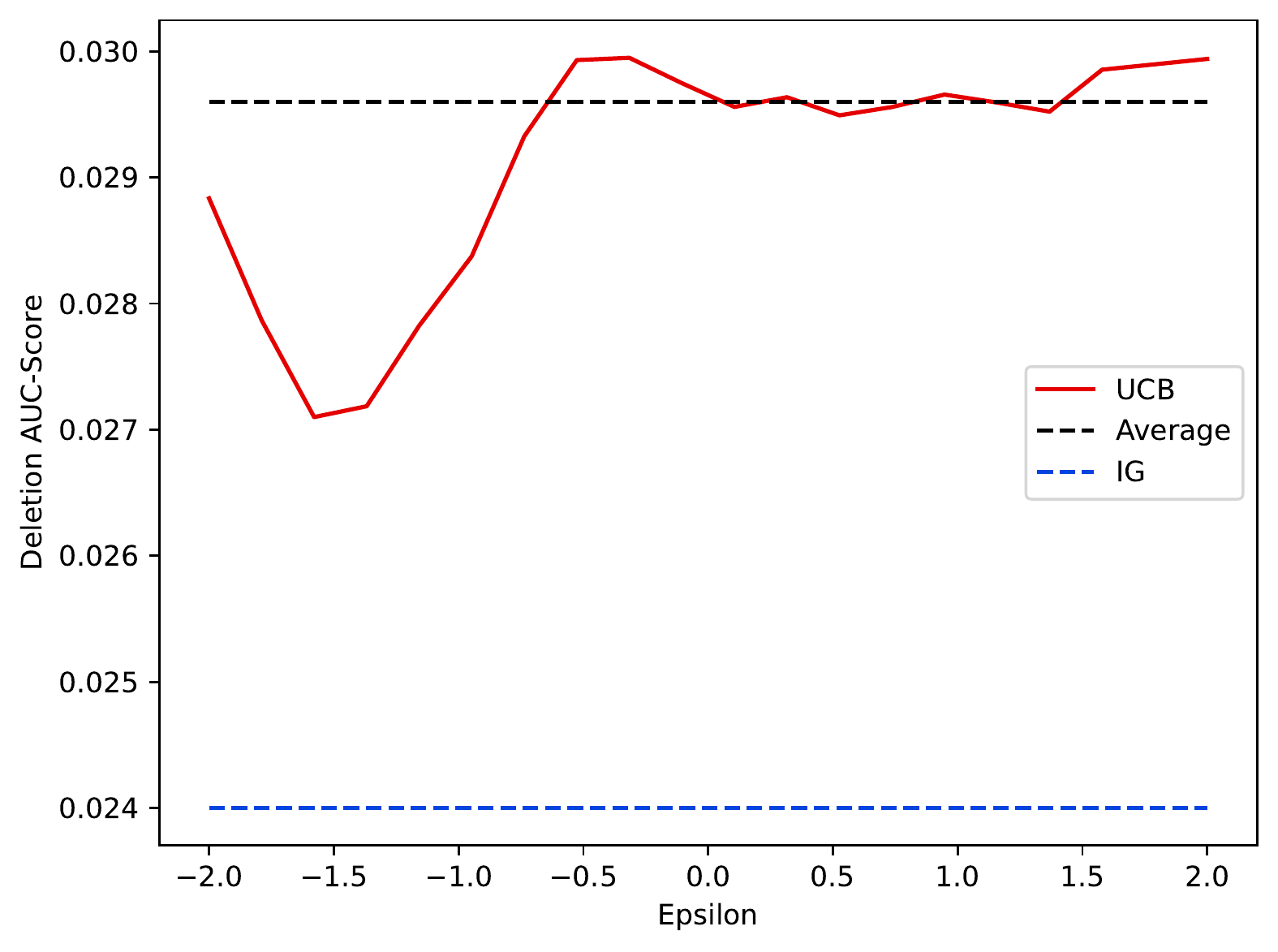}
  \label{fig:ucb_del}
\end{subfigure}
\begin{subfigure}{\textwidth}
  \centering
  \vspace{-.3cm}
  \includegraphics[width=0.9\textwidth]{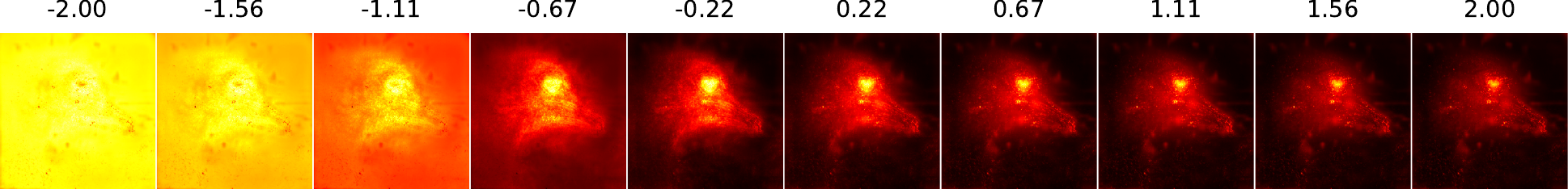}
\end{subfigure}
\caption{Insertion and deletion scores for Upper Confidence Bound (UCB) with varying exploration rate. Base attributions are normalized using Z-score. The explained model is a ResNet18 trained on ImageNet. Results are averaged over 2000 samples.}
\label{fig:ucb_hp}
\end{figure}

For Upper Confidence Bounds, the exploration rate $\epsilon$ controls the balance between mean and standard deviation. An important pixel should score high attributions across all base attributions, such that its mean value is high and standard deviation low. Therefore, we expect that subtracting the standard deviation from the average attribution should increase the performance of the explanation. Figure \ref{fig:ucb_hp} displays the insertion and deletion AUC-score for varying values of $\epsilon \in [-2, 2]$. The experiment confirms our hypothesis as the best insertion score is achieved at $\epsilon=-0.6$. For a large range of $\epsilon$, the insertion score of UCB is higher than the best base attributions and, for a smaller range, better than the average.

The deletion score is not significantly affected by the parameter $\epsilon$ (note the different y-axis-scale), but the best score is achieved at $\epsilon=-1.5$.

\subsubsection{Probability of Improvement}

Probability of Improvement yields an explanation separating important and non-important regions by nearly binary values. This is called an approximated hard explanation and inherently hurts the insertion and deletion score, because the metrics only consider the order of pixel attributions. Nonetheless, sparse and hard masks are more interpretable for humans \cite{yuan20}.

\begin{figure}[ht]
\centering
\begin{subfigure}{.5\textwidth}
  \centering
  \caption{Insertion AUC}
  \includegraphics[width=\linewidth]{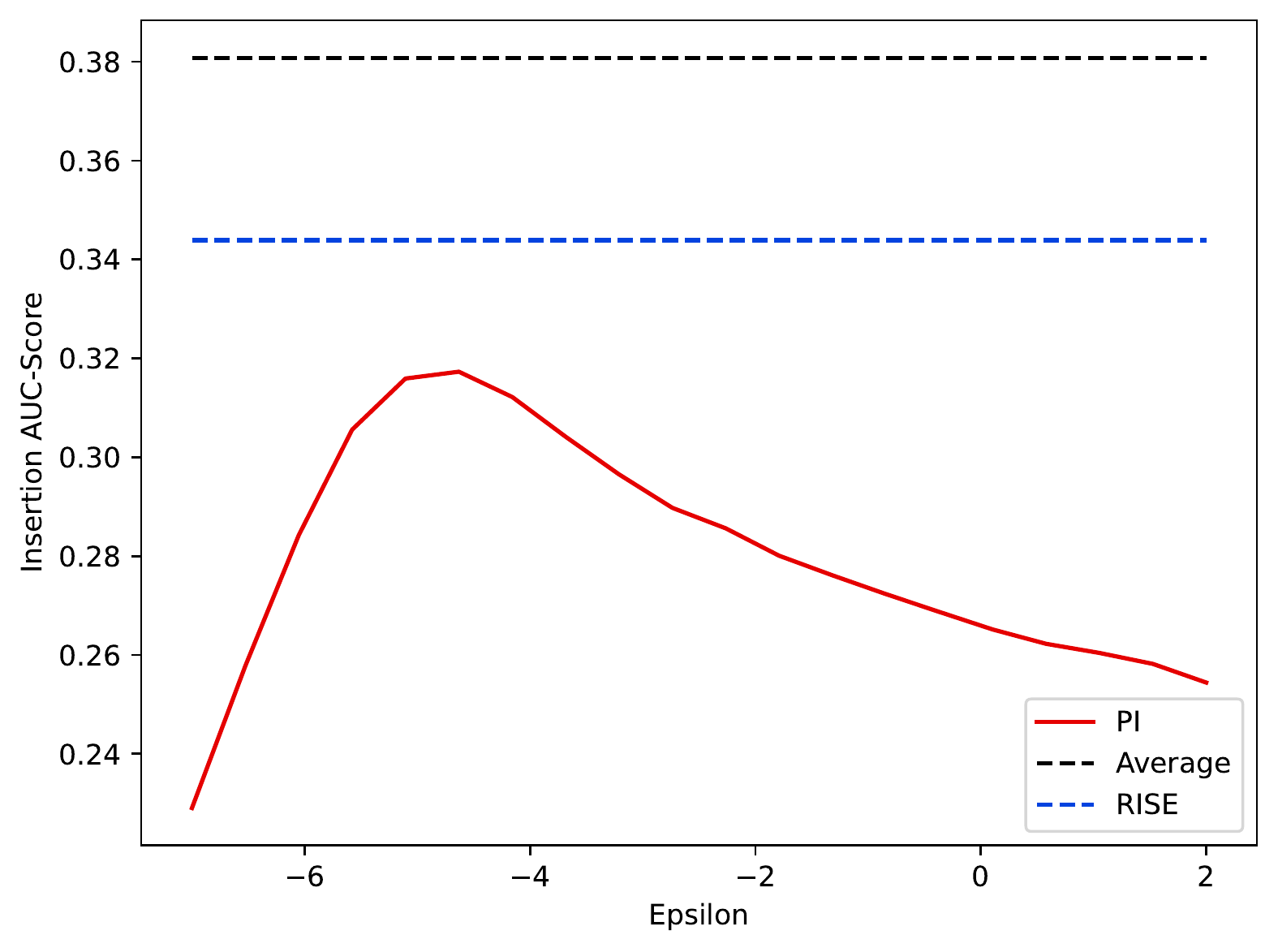}
  \label{fig:pi_ins}
\end{subfigure}%
\begin{subfigure}{.5\textwidth}
  \centering
  \caption{Deletion AUC}
  \includegraphics[width=\linewidth]{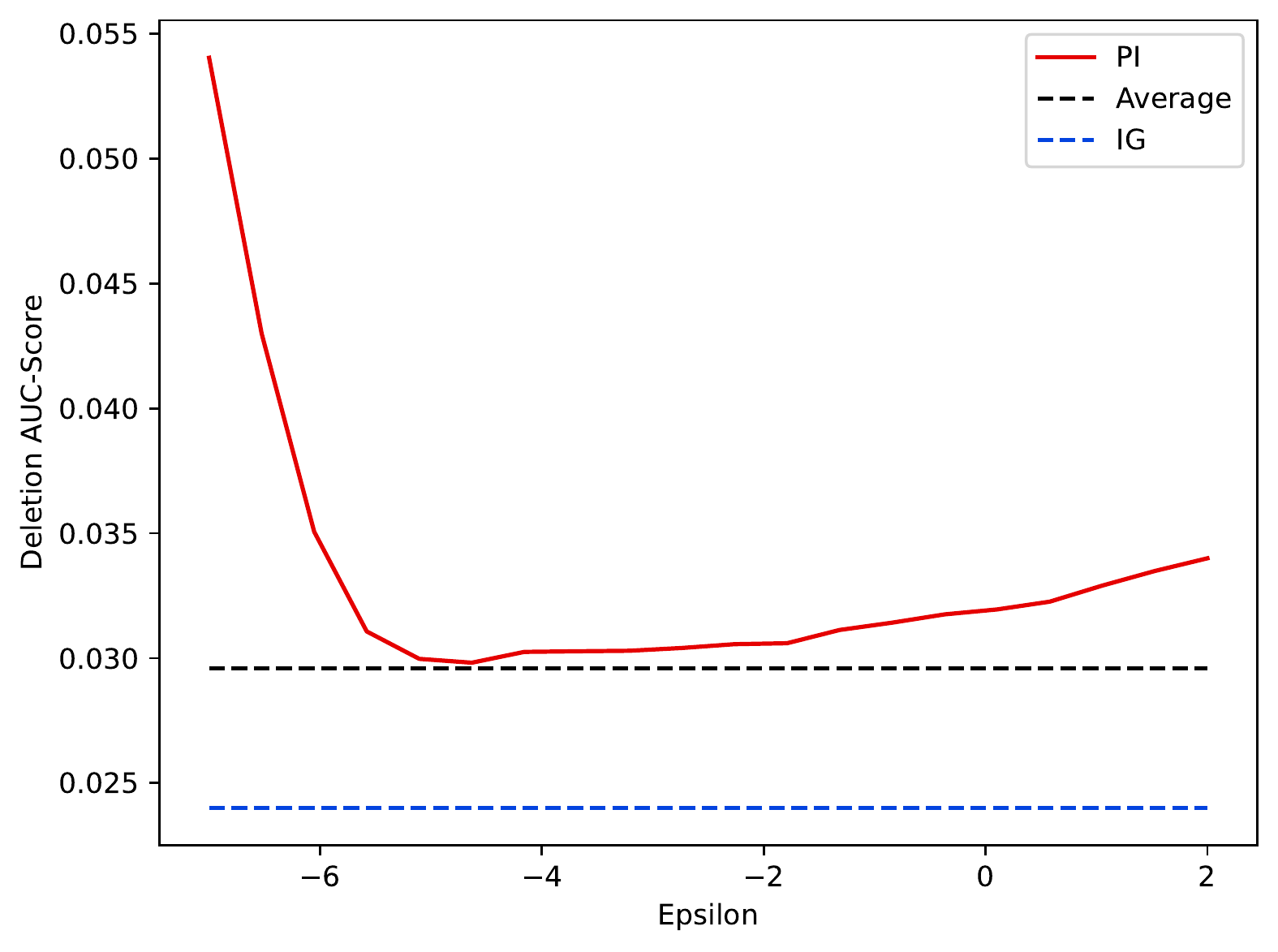}
  \label{fig:pi_del}
\end{subfigure}
\begin{subfigure}{\textwidth}
  \centering
  \vspace{-.3cm}
  \includegraphics[width=0.9\textwidth]{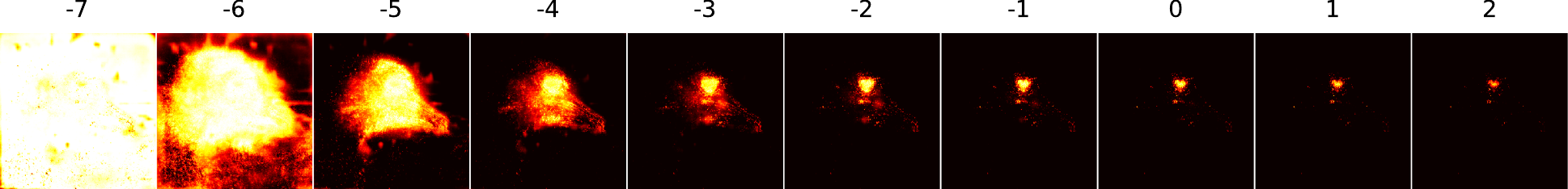}
\end{subfigure}
\caption{Insertion and deletion scores for Probability of Improvement (PI) with varying exploration rate. Base attributions are normalized using Z-score. The explained model is a ResNet18 trained on ImageNet. Results are averaged over 2000 samples.}
\label{fig:pi_hp}
\end{figure}

The exploration rate $\epsilon$ controls the size of the approximated hard saliency map. Figure \ref{fig:pi_hp} visualizes this effect and displays the corresponding insertion and deletion scores. The performance of PI measured by insertion and deletion metric is worse than the best base attribution and average aggregation throughout the whole range. 

\subsubsection{Expected Improvement}

\begin{figure}[ht]
\centering
\begin{subfigure}{.5\textwidth}
  \centering
  \caption{Insertion AUC}
  \includegraphics[width=\linewidth]{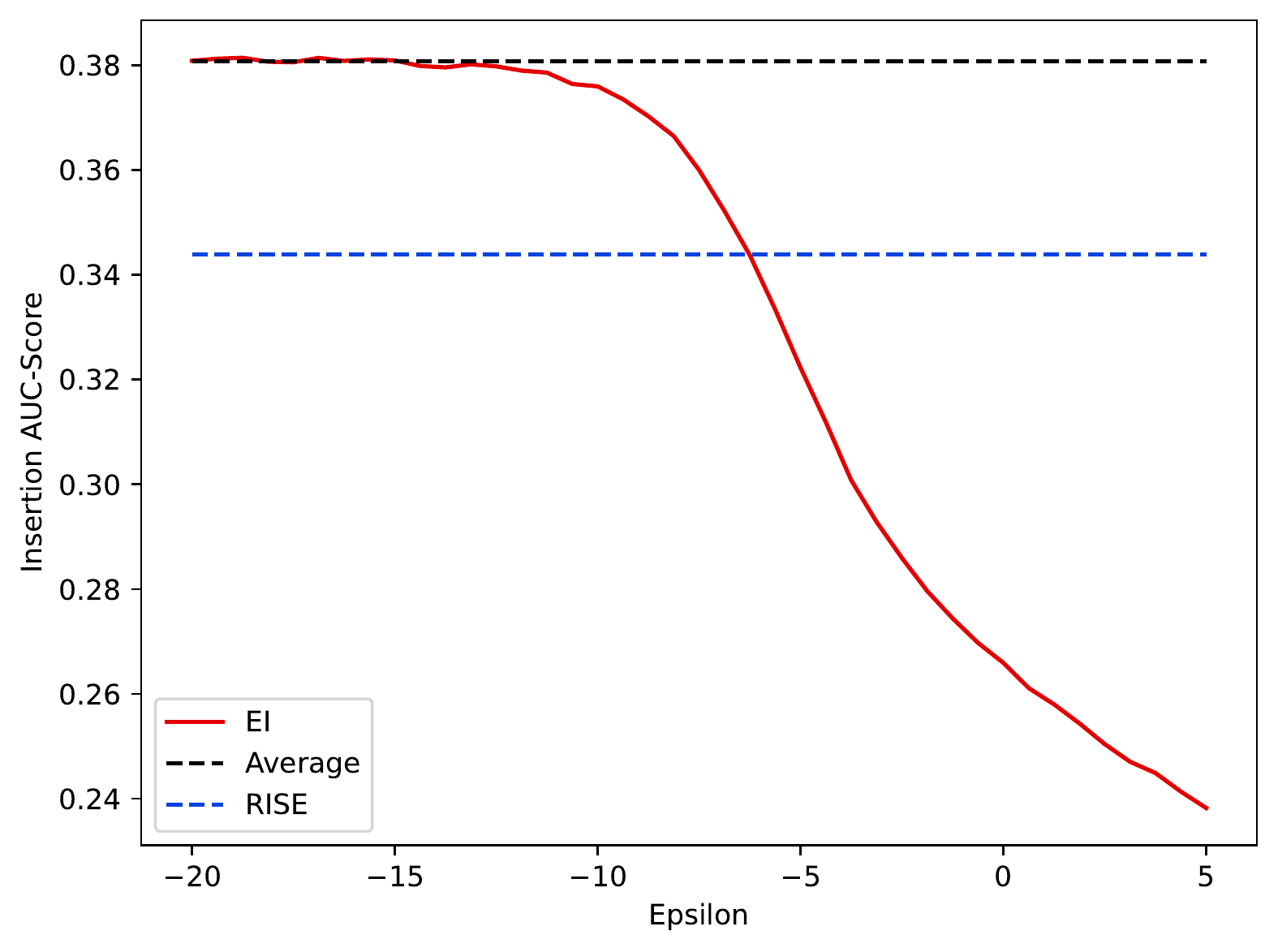}
  \label{fig:ei_ins}
\end{subfigure}%
\begin{subfigure}{.5\textwidth}
  \centering
  \caption{Deletion AUC}
  \includegraphics[width=\linewidth]{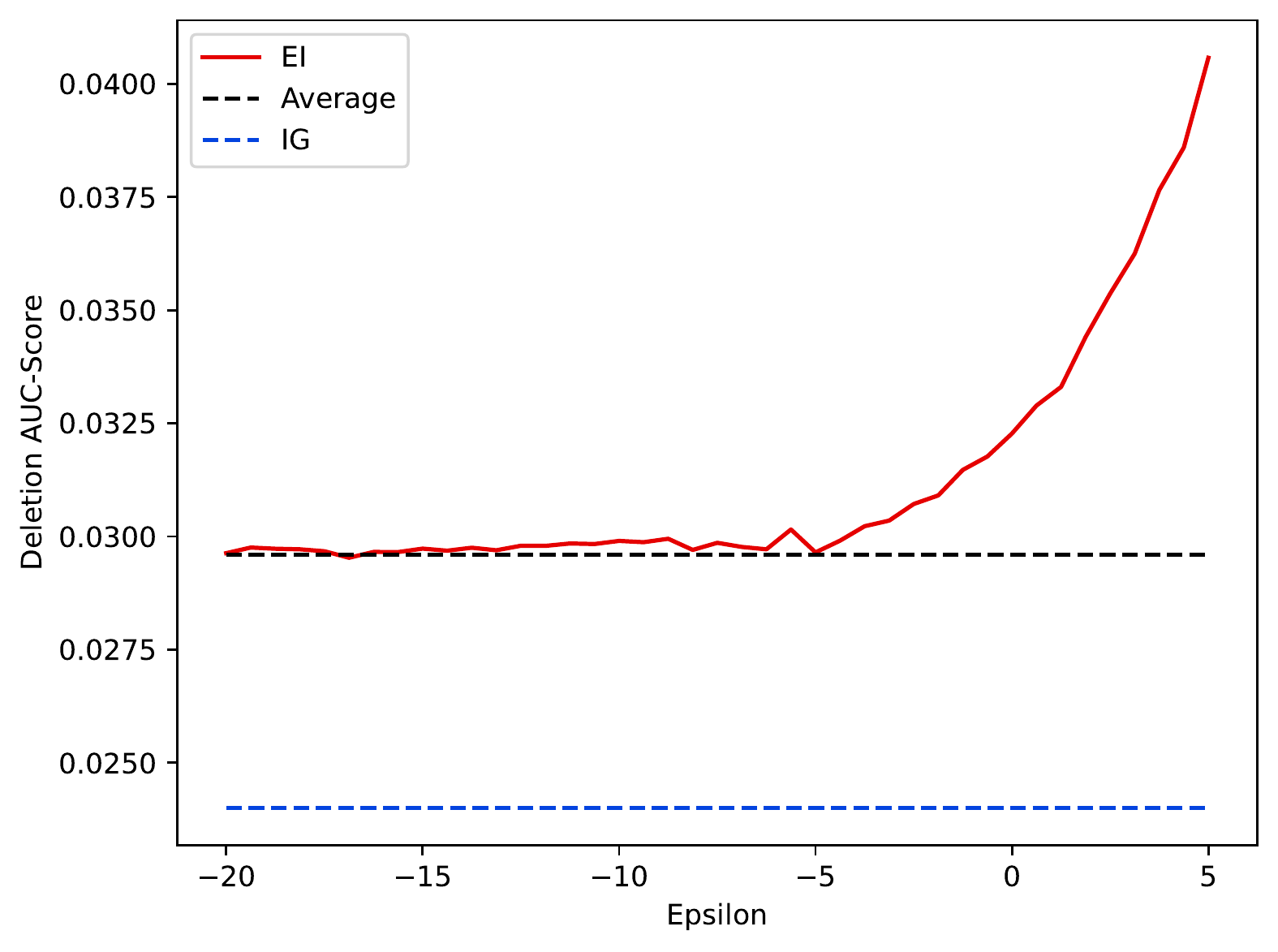}
  \label{fig:ei_del}
\end{subfigure}
\begin{subfigure}{\textwidth}
  \centering
  \vspace{-.3cm}
  \includegraphics[width=0.9\textwidth]{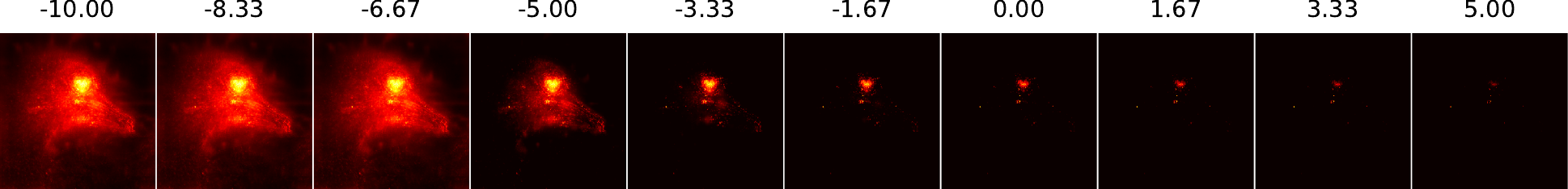}
\end{subfigure}
\caption{Insertion and deletion scores for Expected Improvement (EI) with varying exploration rate. Base attributions are normalized using Z-score. The explained model is a ResNet18 trained on ImageNet. Results are averaged over 2000 samples.}
\label{fig:ei_hp}
\end{figure}

Figure \ref{fig:ei_hp} displays the performance of EI for different exploration rates. For small exploration rate, the performance converges to the average attribution. For larger values, using EI only decreases the performance. Analyzing Equation \ref{eq:ei}, we quickly discover the reason for the convergence to average aggregation. For large negative $\epsilon$, the probability distribution function of a standard normal distribution goes to zero and the cumulative distribution function to one. Therefore, the only terms left are the average aggregation and a constant shift of $- E^+ - \epsilon$. The shift does not change the order of pixel attributions and therefore neither insertion nor deletion score.

\subsubsection{Variance Aggregation}

\begin{figure}[ht]
\centering
\begin{subfigure}{.5\textwidth}
  \centering
  \caption{Insertion AUC}
  \includegraphics[width=\linewidth]{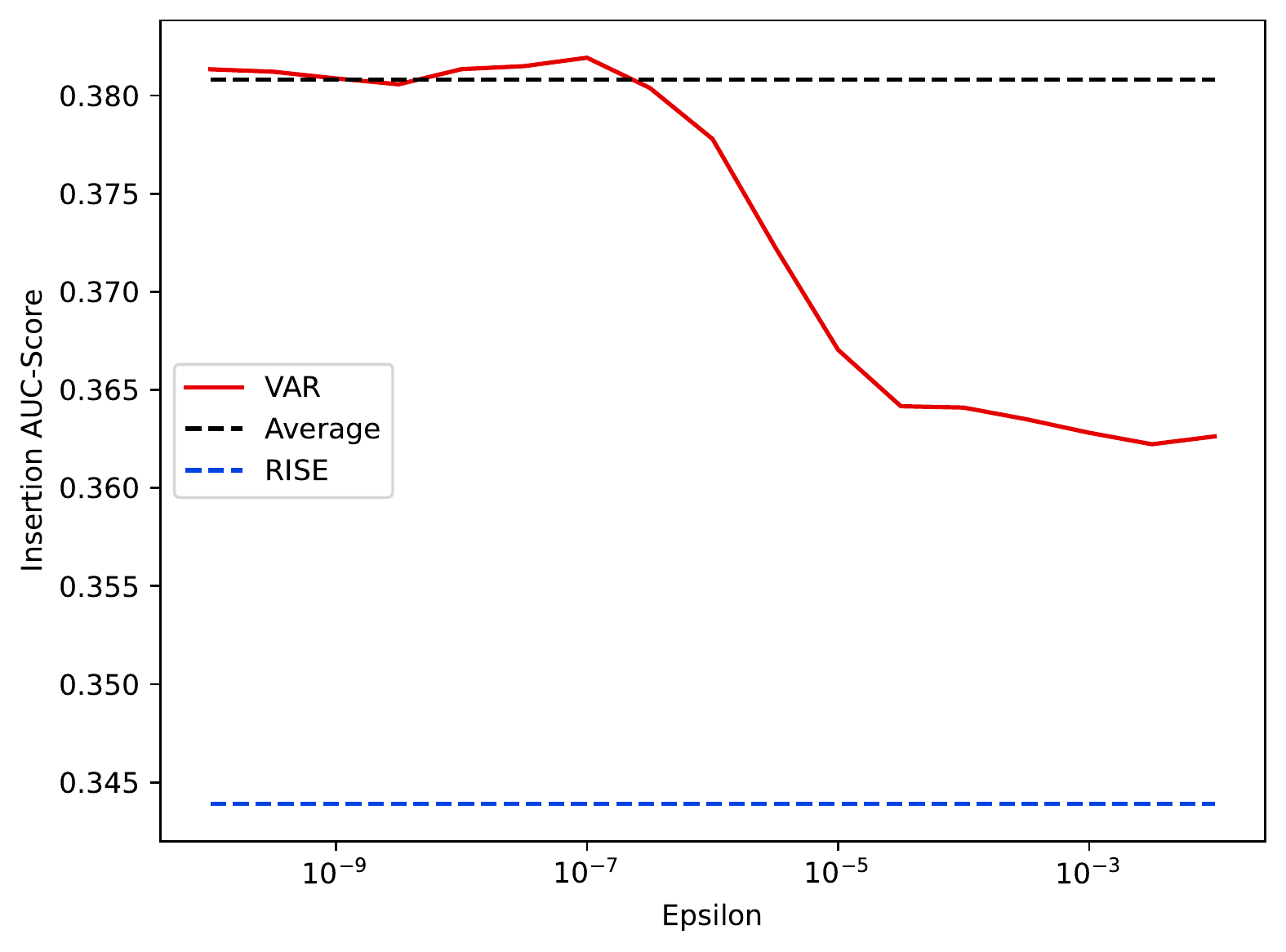}
  \label{fig:var_ins}
\end{subfigure}%
\begin{subfigure}{.5\textwidth}
  \centering
  \caption{Deletion AUC}
  \includegraphics[width=\linewidth]{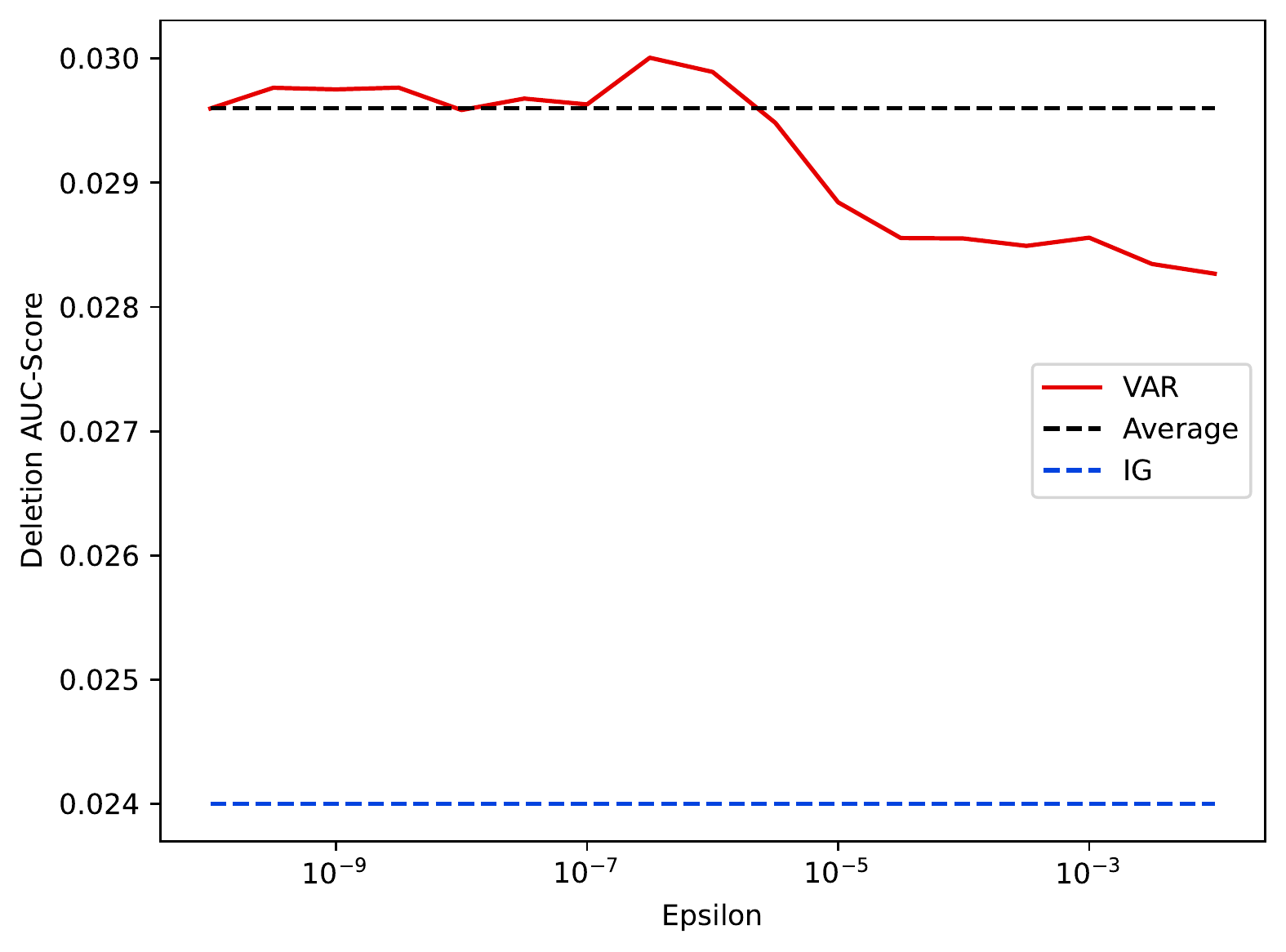}
  \label{fig:var_del}
\end{subfigure}
\begin{subfigure}{\textwidth}
  \centering
  \vspace{-.3cm}
  \includegraphics[width=0.9\textwidth]{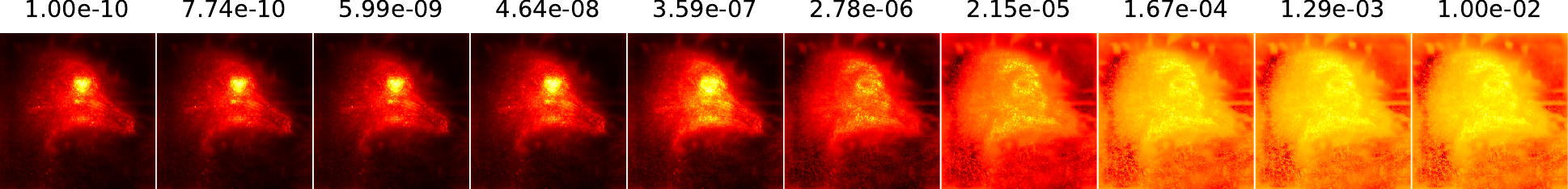}
\end{subfigure}
\caption{Insertion and deletion scores for Variance Aggregation (VAR) with varying exploration rate. Base attributions are normalized using Z-score. The explained model is a ResNet18 trained on ImageNet. Results are averaged over 2000 samples.}
\label{fig:var_hp}
\end{figure}

In Figure \ref{fig:var_hp}, we analyze the fidelity of variance aggregation (VAR) for different values of $\epsilon$. The range of interesting values ($[10^{-10}, 10^{-2}]$) is far from the original paper \cite{rieger19} due to Z-score normalization of the base attributions. For very small values of $\epsilon$, the insertion and deletion score is close to the average attribution. 

Using variance aggregation, it is possible to achieve scores better than the simple average. For insertion, $\epsilon=10^{-7}$ achieves the best result. For deletion, the score becomes better with larger values of $\epsilon$. However, the difference is very small as for all experiments using ImageNet. The best balance is achieved at $\epsilon=10^{-7}$.

\subsection{Effect of Normalization}
\label{sec:exp_norm}

\begin{figure}[ht]
\centering
\begin{subfigure}{.5\textwidth}
  \centering
  \caption{Insertion}
  \includegraphics[width=\linewidth]{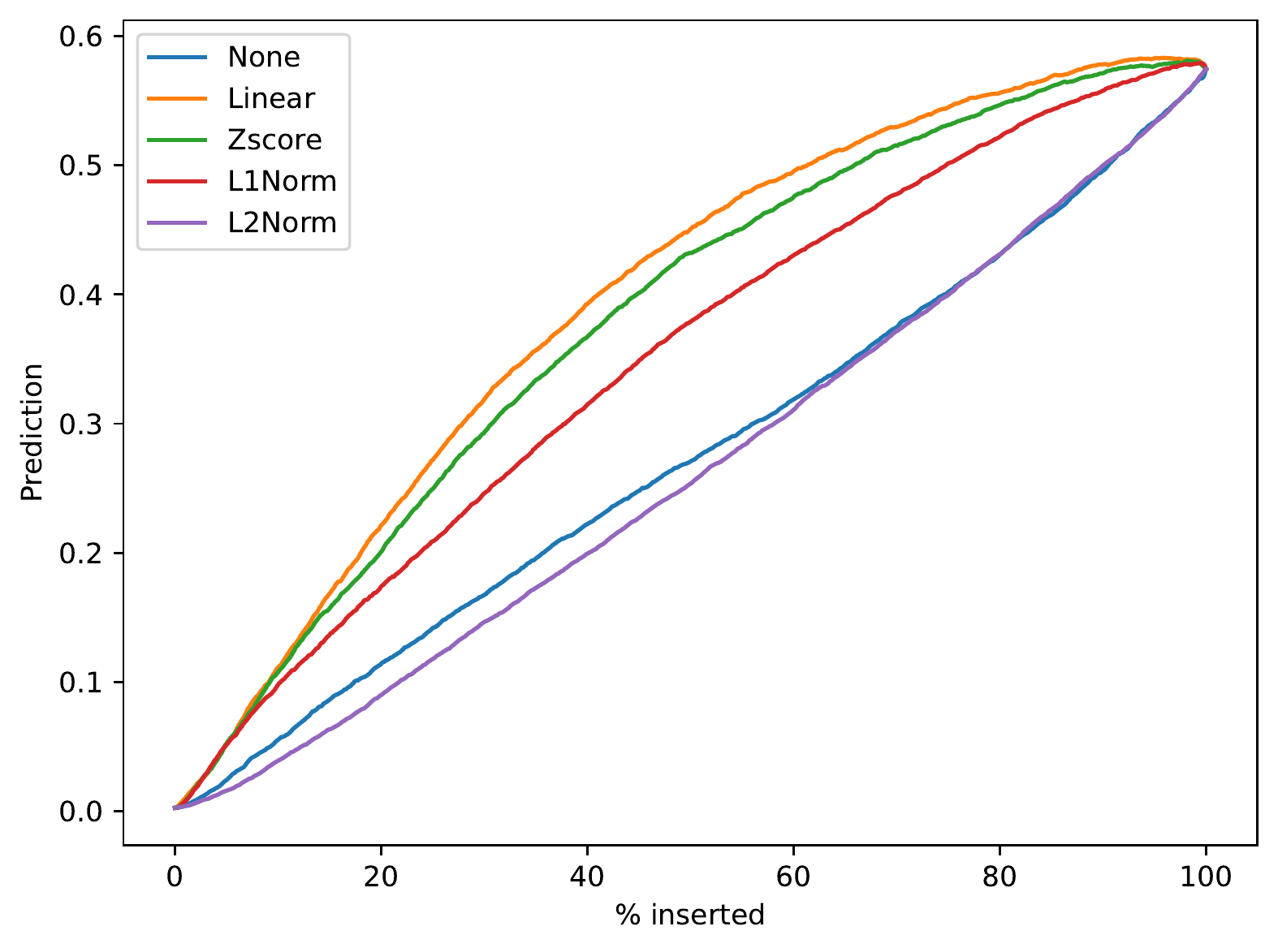}
  \label{fig:norm_ins}
\end{subfigure}%
\begin{subfigure}{.5\textwidth}
  \centering
  \caption{Deletion}
  \includegraphics[width=\linewidth]{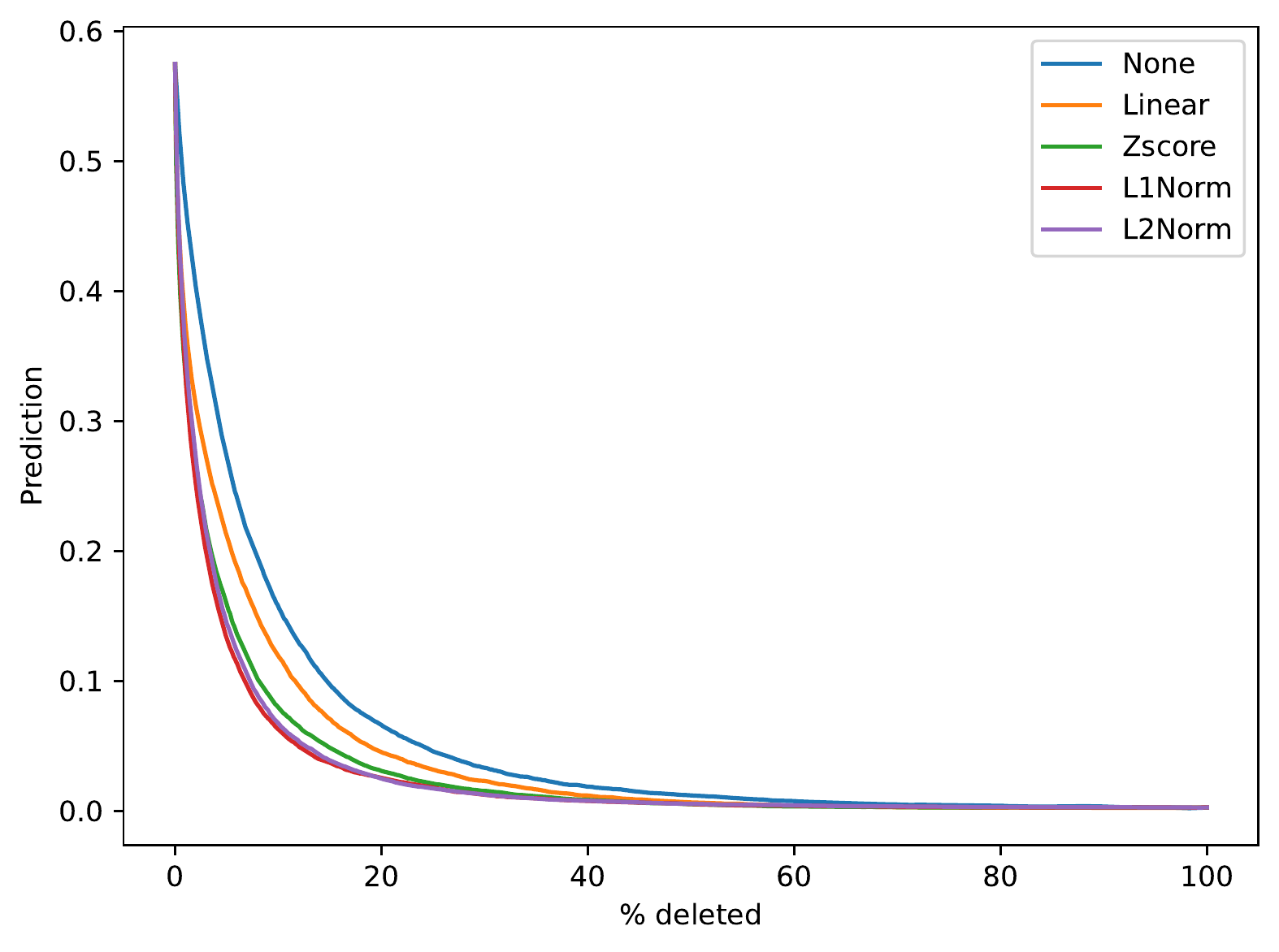}
  \label{fig:norm_del}
\end{subfigure}
\caption{Insertion and deletion curves for different norms using average aggregation. The explained model is a ResNet18 trained on ImageNet. Results are averaged over 2000 samples.}
\label{fig:norm_metric}
\end{figure}

As already discussed above, it is necessary to normalize the base attributions before aggregation. 
However, it is not clear which normalization technique leads to the best result. We compare the aforementioned methods of linear, Z-score, $l_1$- and $l_2$-normalization against no normalization using the average ensemble. 
Figure \ref{fig:norm_metric} displays the insertion and deletion curves for this experimental setup. Interestingly, linear normalization achieves, despite its theoretical downside of ignoring the full distribution, the best insertion score. It is closely followed by Z-score normalization. 
The $l_1$- and $l_2$-normalization achieve the best deletion score, again closely followed by Z-score normalization. 
We conclude that Z-score normalization leads to the best balance between a good insertion and deletion score, even though it does not perform best in either metric. 

\subsection{Ablation Study}
An important question is the sensitivity of the aggregation methods to unfaithful or noisy base attributions. Borisov et al.\ \cite{borisov21} included multiple base attributions consisting of only random noise and visually compared the results. However, this experiment does not reflect reality well. We focus on the question if the worst base attributions of an ensemble contribute positively to the aggregated result.

\begin{figure}[ht]
\centering
\begin{subfigure}{.5\textwidth}
  \centering
  \includegraphics[width=\linewidth]{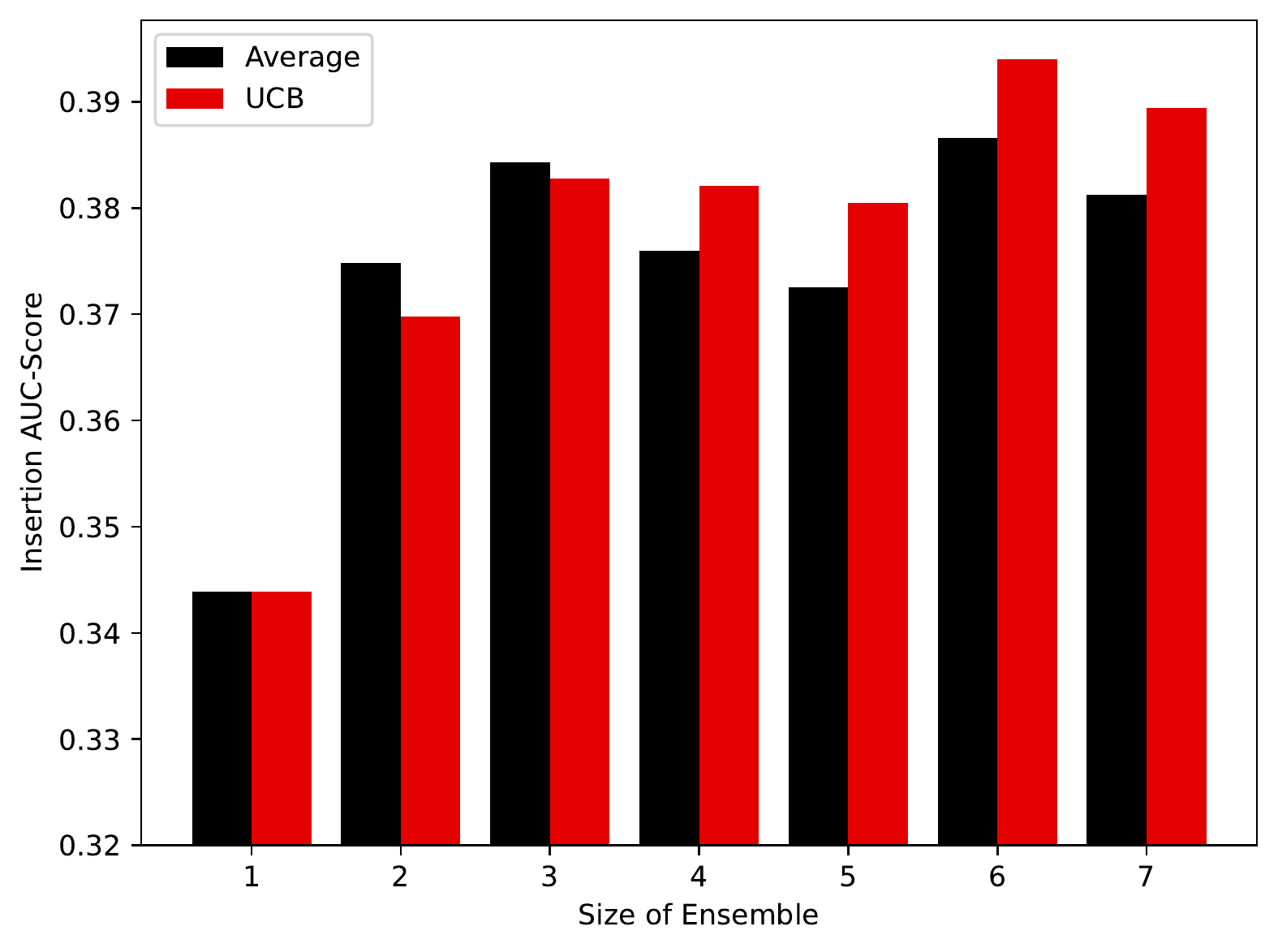}
  \caption{Insertion}
  \label{fig:abl_ins}
\end{subfigure}%
\begin{subfigure}{.5\textwidth}
  \centering
  \includegraphics[width=\linewidth]{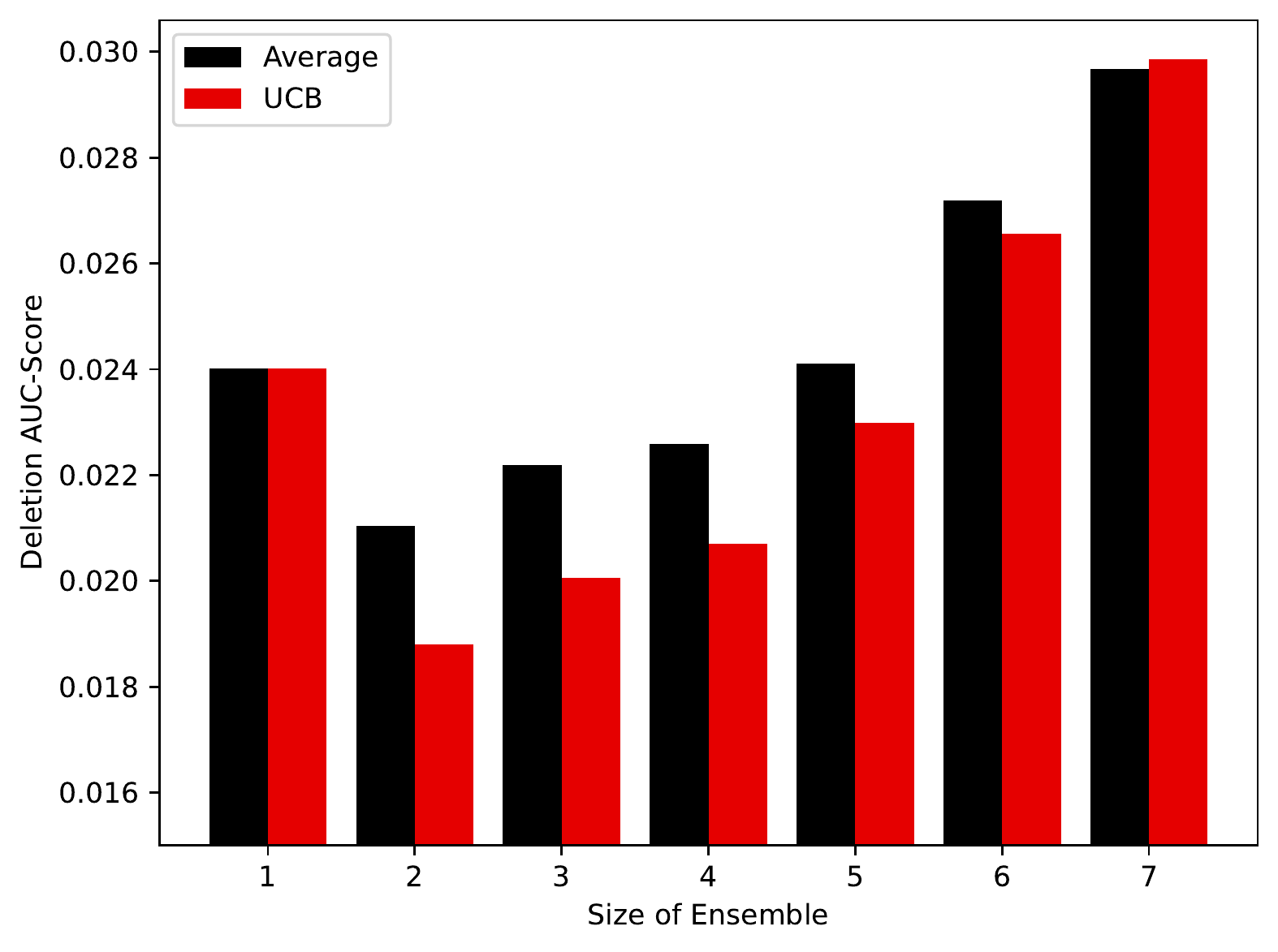}
  \caption{Deletion}
  \label{fig:abl_del}
\end{subfigure}
\caption{Insertion and deletion for different ensemble sizes. Ensembles contain the top-k base attributions regarding the measured metric. The explained model is a ResNet18 trained on ImageNet. Results are averaged over 2000 samples.}
\label{fig:abl}
\end{figure}

To answer this question, we start with the best base attribution and iteratively insert the other base attributions in descending order of their performance. We consider the performance regarding both metrics individually, such that the order is different for each metric. Figure \ref{fig:abl} displays the results of this experiment. Using the insertion metric, we observe a negative impact when inserting the fourth, fifth and seventh base attribution into the ensemble. This indicates that better performance is possible if fewer, high performing base attributions are used. However, the drop in performance is less pronounced in UCB, compared to average attribution, due to the inclusion of the standard deviation between base attributions.

Again, the order of magnitude for the insertion score is much smaller (note the different y-axis-scale), but best performance is achieved when only using two ensembles. Again, for smaller ensembles average attribution performs better than UCB, but for larger ensembles the opposite is true.

Even though performance decreases for both metrics when including weaker base attributions, the drop in performance is relatively small. In a real-world scenario, determining the performance of every base attribution and finding the best ensemble would be a tedious task. Therefore, we argue that using all available base attributions and aggregating the ensemble with UCB is the best option. 

Another interesting observation regarding both metrics is the increasing difference in performance between UCB and the simple average when using larger ensembles. This makes sense, as the standard deviation between base attributions becomes more meaningful in larger ensembles. 

\FloatBarrier
\section{Conclusion and Future Work}
We introduced multiple aggregation functions originating in the context of Bayesian Optimization and compared their fidelity to existing aggregation functions. The empirical comparison revealed that incorporating confidence using Upper Confidence Bounds yields the best saliency maps. 

Additionally, normalization of the base attributions is very important. We compared multiple normalization techniques and found Z-score normalization most effective. 

As discussed above, aggregation functions like Expected Improvement and Probability of Improvement assign an attribution of zero to many pixels, which inherently hurts the insertion and deletion scores. We tried to solve this issue along with the problem of finding the best exploration rate by averaging over multiple exploration rates. This solution only achieved limited results, but in future work we may examine other techniques. For example, one could successively mask the most important pixels and compute new attribution scores in every iteration. In addition to solving the problem of attribution scores of zero, this solution also models the sequential process of Bayesian Optimization. Therefore, it may yield a better theoretical foundation and empirical results. However, this is beyond the scope of this work.

Many base attributions have a high number of hyperparameters, which significantly affect the resulting explanation. For example, iGOS++ \cite{khorram20} has 12 hyperparameters. Finding the best hyperparameter configuration can be a difficult challenge \cite{feurer19}. As discussed by Borisov et al.\ \cite{borisov21}, aggregating explanations from multiple hyperparameter configurations alleviates this issue and might even produce better results. Further directions of research may investigate the applicability of the presented methods in this context. 

Furthermore, all aggregation methods presented are pixel-level aggregation, i.e. the aggregated attribution is calculated independently for each pixel. The only exception is RBM, which can be seen as a global aggregation since it considers all pixels during training time. Future work could also explore aggregations based on a local neighborhood.

\FloatBarrier
% \bibliographystyle{plain}
% \bibliography{references}
\newpage
\printbibliography

\newpage
\appendix
\section{Hyperparameter of Base Attributions}
\label{sec:hp_conf}
The hyperparameters of the base attributions used for all ensembles in the experiments are listed below:
\begin{itemize}
\setlength\itemsep{-0.1cm}
    \item SmoothGrad \cite{smilkov17}
    \vspace{-.2cm}
    \begin{itemize}
        \item 256 samples
        \item Standard deviation $\sigma = 0.1$
    \end{itemize}
    
    \item Integrated Gradients \cite{sundararajan17} with blur baseline ($\sigma = 20$ and kernel size $25\times25$)
    \vspace{-.2cm}
    \begin{itemize}
        \item Baseline: blurred image with $\sigma=20$ and kernel size $25\times25$
        \item 30 interpolation steps
    \end{itemize}
    
    \item Expected Gradients \cite{erion19}
    \vspace{-.2cm}
    \begin{itemize}
        \item 20 samples
        \item 30 interpolation steps
        \item Baseline: Normal
    \end{itemize}
    
    \item GradCAM \cite{selvaraju19} (has no parameters)
    
    \item PolyCAM \cite{englebert22} 
    \vspace{-.2cm}
    \begin{itemize}
        \item Channel-wise variation of confidence as activation map weights (see original paper for more information)
    \end{itemize}
    
    \item iGOS++ \cite{khorram20}
    \vspace{-.2cm}
    \begin{itemize}
        \item Mask resolution $50\times50$
        \item Sparsity regularization parameter $\lambda_1=20$
        \item Smoothness regularization parameter $\lambda_2=30$
        \item 20 iterations
        \item 25 interpolation steps
        \item Maximum step size $\alpha_u = 10$
        \item Minimum step size $\alpha_l = 10^{-5}$
        \item Armijo condition constant $\beta = 2$
        \item Bilateral Total Variation Norm $\beta = 10^{-4}$
        \item Bilateral Total Variation Smoothing $\sigma = 0.01$
        \item decay $\eta = 0.2$
        \item Baseline: Normal
    \end{itemize}
    
    \item RISE \cite{petsiuk18}
    \vspace{-.2cm}
    \begin{itemize}
        \item 5000 samples
        \item Mask resolution $10\times10$
    \end{itemize}
\end{itemize}

Normal baseline refers to an estimation of mean and standard deviation of each color channel. Afterwards, new pixel values are sampled from the resulting normal distribution.

\end{document}